%% file: main.tex
\documentclass[10pt,twocolumn,letterpaper]{article}

\usepackage{iccv}
\usepackage{times}
\usepackage{epsfig}
\usepackage{graphicx}
\usepackage{amsmath}
\usepackage{amssymb}
\usepackage[table]{xcolor}

\usepackage{graphicx}
\usepackage{booktabs}
\usepackage{multirow}
\usepackage{enumitem}
\usepackage{array}
\usepackage{makecell}
\usepackage{microtype}
\usepackage{caption}
\usepackage{subcaption}
\usepackage{circledsteps}
\usepackage{color, colortbl}
\usepackage[pagebackref=true,breaklinks=true,letterpaper=true,colorlinks,bookmarks=false]{hyperref}

\usepackage{fancyhdr}

\usepackage[capitalize]{cleveref}
\crefname{section}{Sec.}{Secs.}
\Crefname{section}{Section}{Sections}
\Crefname{table}{Table}{Tables}
\crefname{table}{Tab.}{Tabs.}
\usepackage{booktabs,caption}
\iccvfinalcopy % *** Uncomment this line for the final submission

\def\iccvPaperID{6595} % *** Enter the ICCV Paper ID here

\def\Approach{GaPro}
\def\Problem{BS-3DIS}

\definecolor{mydarkblue}{rgb}{0,0.08,1}
\definecolor{mydarkgreen}{rgb}{0.02,0.6,0.02}
\definecolor{myred}{rgb}{1.0,0.0,0.0}

\ificcvfinal\fi

\begin{document}
\input{definitions}
%%%%%%%%% TITLE
\title{\Approach: Box-Supervised 3D Point Cloud Instance Segmentation \\Using Gaussian Processes as Pseudo Labelers}

\author{Tuan Duc Ngo \qquad Binh-Son Hua \qquad Khoi Nguyen\\
VinAI Research, Hanoi, Vietnam\\
{\tt\small \{v.tuannd42, v.sonhb, v.khoindm\}@vinai.io}
}

\maketitle

\thispagestyle{plain}
\pagestyle{plain}

%%%%%%%%% ABSTRACT
\begin{abstract}
\vspace{-10pt}
   Instance segmentation on 3D point clouds (3DIS) is a longstanding challenge in computer vision, where state-of-the-art methods are mainly based on full supervision. As annotating ground truth dense instance masks is tedious and expensive, solving 3DIS with weak supervision has become more practical. In this paper, we propose \Approach, a new instance segmentation for 3D point clouds using axis-aligned 3D bounding box supervision. 
   Our two-step approach involves generating pseudo labels from box annotations and training a 3DIS network with the resulting labels. Additionally, we employ the self-training strategy to improve the performance of our method further. 
   We devise an effective Gaussian Process to generate pseudo instance masks from the bounding boxes and resolve ambiguities when they overlap, resulting in pseudo instance masks with their uncertainty values. Our experiments show that \Approach~outperforms previous weakly supervised 3D instance segmentation methods and has competitive performance compared to state-of-the-art fully supervised ones. Furthermore, we demonstrate the robustness of our approach, where we can adapt various state-of-the-art fully supervised methods to the weak supervision task by using our pseudo labels for training. 
   The source code and trained models are available at \url{https://github.com/VinAIResearch/GaPro}.   
\end{abstract}

\begin{figure}[t]
  \centering
  \includegraphics[width=1.0\linewidth]{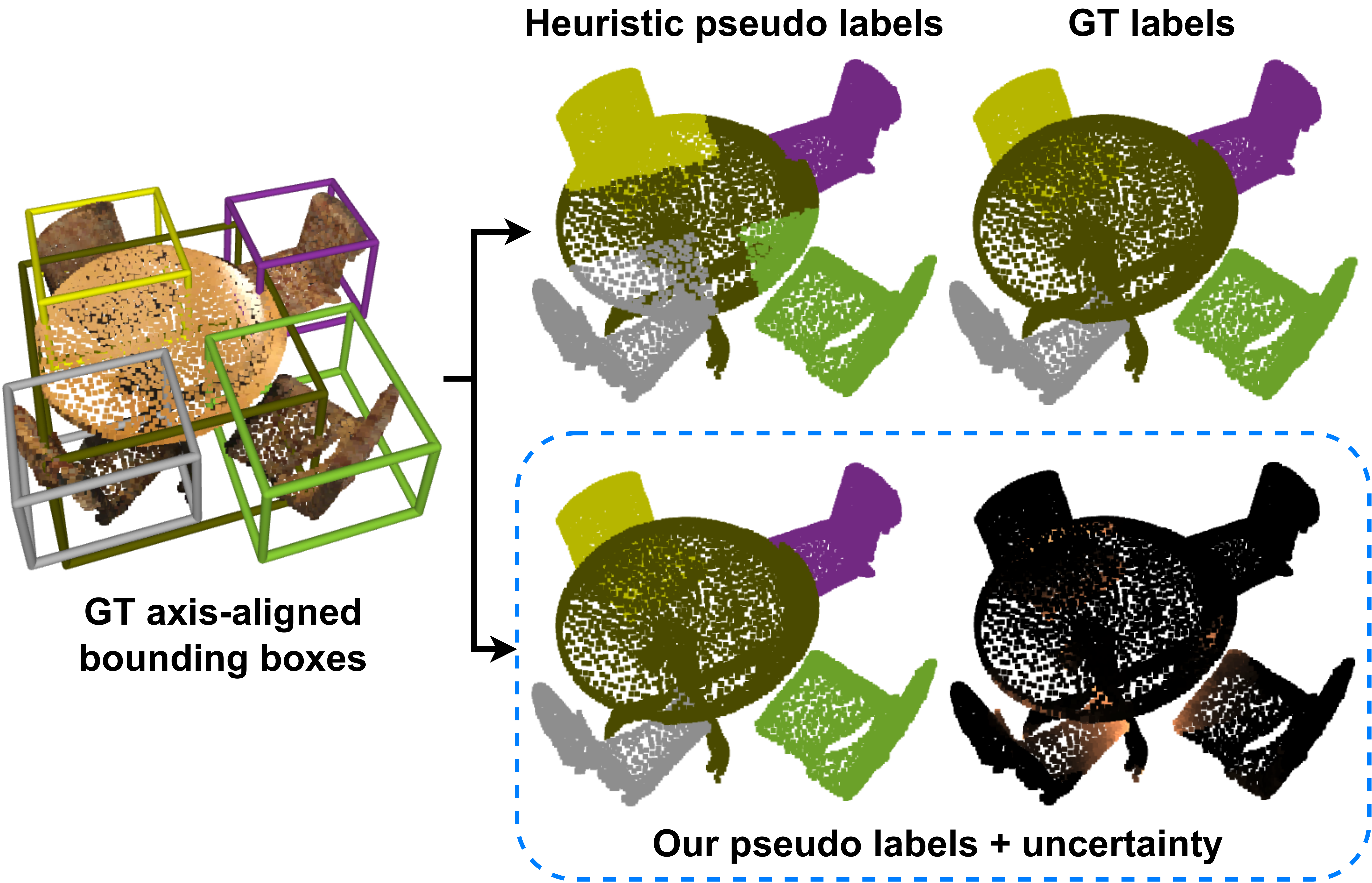}
   % \vspace{-4pt}
   \caption{Weakly supervised instance segmentation relies on high-quality pseudo labels to achieve competitive performance. Given only axis-aligned bounding box annotations, pseudo labels based on heuristics~\cite{chibane2021box2mask} often have large errors in box overlapping regions and thus yield inferior performance. Our \Approach~predicts the pseudo labels and their confidence using Gaussian Processes, via resolving the ambiguity in box overlapping regions.
   }
   \label{fig:teaser}
   \vspace{-16pt}
\end{figure}

%%%%%%%%% BODY TEXT
\vspace{-13pt}
\section{Introduction}
This paper addresses the challenging problem of box-supervised 3D point cloud instance segmentation (\Problem), which seeks to segment every point into instances of predefined classes using only axis-aligned 3D bounding boxes as supervision during training.
This problem arises to address the huge annotating cost of fully-supervised 3D point cloud instance segmentation (3DIS) where every point in the point cloud is manually labeled.
Compared to 3DIS, \Problem~is considered significantly harder. First, axis-aligned boxes cannot capture the shape or geometry of objects as they only represent the very coarse extent of the objects. Second, unlike instance mask where points only belong to at most one mask, points can belong to multiple boxes as visualized in Fig.~\ref{fig:teaser}, resulting in the ambiguous point-object assignment.

The task of box-supervised 3D point cloud instance segmentation has received little attention, with Box2Mask \cite{chibane2021box2mask} being the first attempt. However, due to the ambiguity in point-object assignments, the point-wise predicted boxes are unreliable for clustering. This leads to a significant performance gap compared to fully supervised methods, such as Mask3D \cite{Schult23ICRA}, which achieves an mAP of 55.2 on ScanNetV2 \cite{dai2017scannet}, compared to Box2Mask's 39.1 (around 30\%) using the same backbone. Furthermore, Box2Mask is not adaptable to new advances in fully supervised 3DIS, as it is designed as a standalone method.

To address these limitations, we propose a novel pseudo-labeling method that can be used as a universal plugin for any 3DIS network and offers an instant solution for any new fully supervised 3DIS approach, with a smaller performance gap between fully supervised and \Problem~versions, typically around 10\%.
In particular, we formulate it as a learning problem with two unknowns: the network's parameters and the ground-truth object masks. Our goal is to construct pseudo object masks from box supervision and optimize the network's parameters using these pseudo labels. To achieve this, we propose using Gaussian Process (GP) on each pair of overlapping 3D bounding boxes to infer the optimal pseudo labels of object masks and their uncertainty values, which are constrained by the given 3D bounding boxes. Next, we modify a 3DIS network to predict additional uncertainty values along with the object mask to match the inferred pseudo labels obtained from the GP.
GP plays a key role in our approach. First, it models the similarity relationship among regions of the point cloud, which enables effective label propagation from determined regions (belonging to a single box) to undetermined regions (belonging to multiple boxes). Second, it estimates the uncertainty of the predictions with weak labels, providing informative indications for annotators to correct uncertain regions of pseudo labels for training the 3D instance segmentation network.

We evaluate our approach on various state-of-the-art 3DIS methods, including PointGroup \cite{jiang2020pointgroup}, SSTNet \cite{liang2021instance}, SoftGroup \cite{vu2022softgroup}, ISBNet~\cite{ngo2023isbnet}, and SPFormer \cite{sun2022superpoint}, using two challenging datasets: ScanNetV2 \cite{dai2017scannet} and S3DIS \cite{armeni2017joint}. Our box-supervised versions of these methods achieve comparable performance to their fully-supervised counterparts on both datasets, outperforming other weakly-supervised 3DIS methods significantly.

In summary, the contributions of our work are as follows:

\begin{itemize}[noitemsep,topsep=0pt]
    \item We propose \Approach, a weakly-supervised 3DIS method based on 3D bounding box supervision. We devise a systematic approach to generate pseudo object masks from 3D axis-aligned bounding boxes so that fully supervised 3DIS methods can be retargeted for weak supervision purposes. 
    \item We propose an efficient Gaussian Process to resolve the ambiguity of pseudo labels in the overlapped region of two or more bounding boxes by inferring both the pseudo masks and their uncertainty values. 
    \item Our \Approach~achieves competitive performance with the SOTA fully-supervised approaches and outperforms other weakly-supervised methods by a large margin on both ScanNetV2 and S3DIS datasets.
\end{itemize}

In the following, Sec.~\ref{sec:related_work} reviews prior work; Sec.~\ref{sec:approach} specifies~\Approach; and Sec.~\ref{sec:experiments} presents our implementation details and experimental results. Sec.~\ref{sec:conclusion} concludes with some remarks and discussions.

\begin{figure*}[t]
  \centering
  \includegraphics[width=0.88\linewidth]{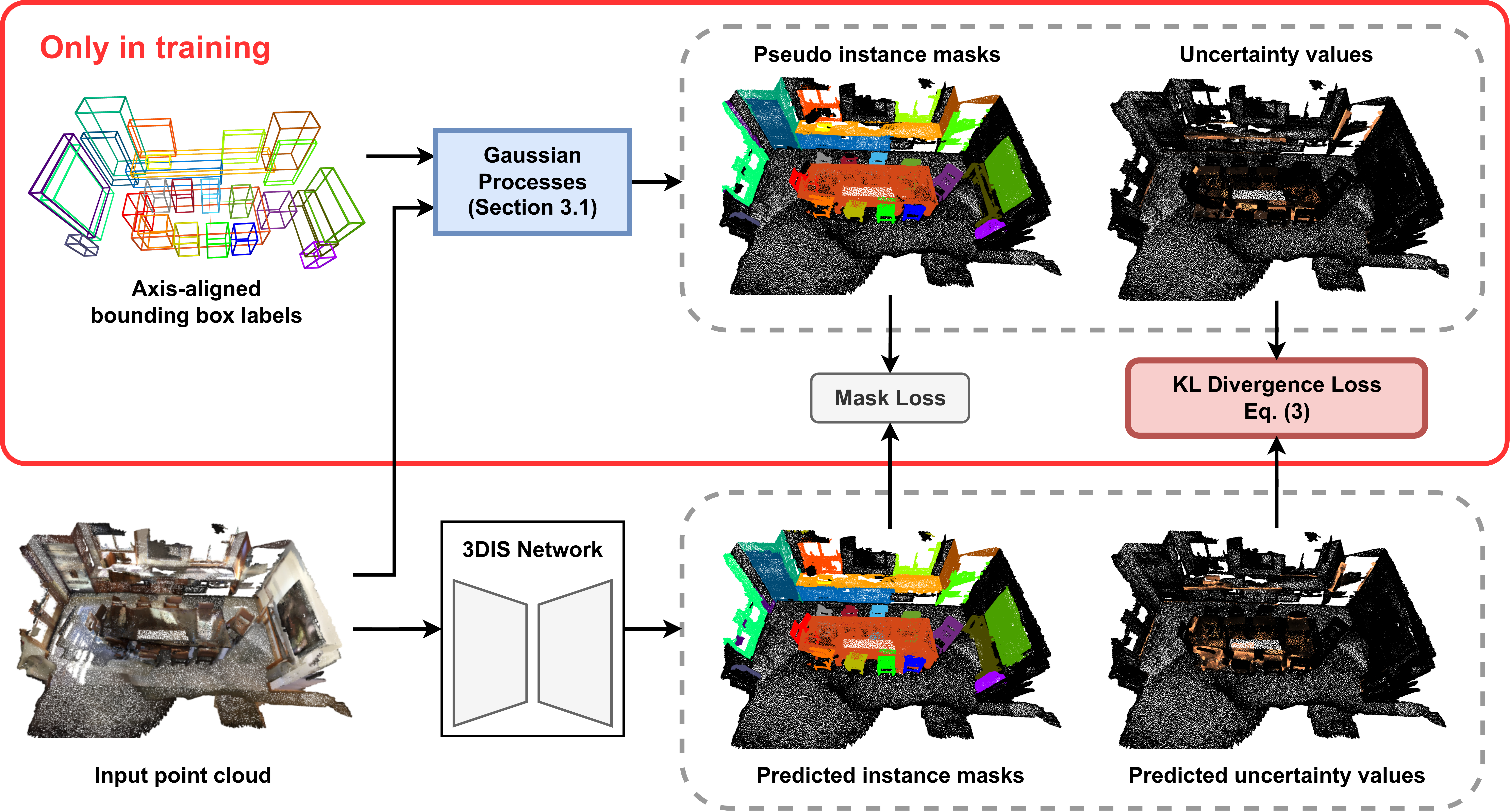}
   % \vspace{-4pt}
   \caption{\textbf{Overall architecture of our approach}. \Approach~is a two-step approach consisting of leveraging Gaussian Processes to generate pseudo instance masks and their uncertainty values, and training a 3DIS network to match its prediction against these pseudo labels with a new KL divergence loss along with the mask loss. 
   }
   \label{fig:architecture}
   \vspace{-12pt}
\end{figure*}

\section{Related Work}
\label{sec:related_work}

This section reviews some related work on 3D point cloud instance segmentation and weakly-supervised instance segmentation in 2D and 3D, and the usage of the Gaussian Process in the 3D point cloud. 

\myheading{3D Point Cloud Instance Segmentation (3DIS)} approaches are categorized into box-based, cluster-based, and dynamic convolution (DC)-based methods. Box-based methods \cite{hou20193d,yang2019learning,yi2019gspn} detect and segment the foreground region inside each 3D proposal box to get instance masks. Cluster-based methods cluster points into instances based on the predicted object centroid \cite{wang2018sgpn,jiang2020pointgroup,chen2021hierarchical,vu2022softgroup,dong2022learning}, or build a tree/graph then cut the subtrees/subgraphs as clusters \cite{liang2021instance,hui2022learning}. DC-based methods \cite{he2021dyco3d,sun2022superpoint,he2022pointinst3d,wu20223d,Schult23ICRA,liu20223d} generate kernels representing different object instances to convolve with point-wise features to produce instance masks.
Among these methods, DC-based approaches are preferred due to their superior performance, since they do not rely on error-prone intermediate predictions like proposal boxes or clusters. 
However, fully-supervised 3DIS approaches require costly point-wise instance annotation for training which hinders their application in practice. Our proposed approach only uses 3D instance boxes (represented by two points) as supervision, which is much cheaper to obtain. Our approach can be applied to all the aforementioned fully-supervised 3DIS approaches, allowing them to transform into \Problem~versions.

\myheading{Weakly-supervised 2D image instance segmentation} aims to segment images into instances of predefined classes using weaker supervision than instance masks. Different types of weak supervision include image-level classes \cite{Durand2017WILDCATWS, laradji2019masks, zhou2018weakly}, instance points \cite{cheng2022pointly,tang2022active,fan2022pointly}, and instance boxes \cite{hsu2019weakly, tian2020boxinst, zhang2021affinity, lee2021bbam, lan2021discobox, li2022box, cheng2022boxteacher, yang2022asyinst, li2022box2mask, lan2023vision}. Box supervision is particularly attractive because it provides a stronger signal for training with only two points per instance. Box-supervised approaches (BS-2DIS) compensate for the lack of ground-truth masks by regularizing the training of instance segmenters with priors.
Various methods have been proposed for BS-2DIS, such as BoxInst \cite{tian2020boxinst} with tight-box prior loss and color smoothness, LevelSetBox \cite{li2022box} with level set evolution, Mask Auto-Labelers \cite{lan2023vision} using Conditional Random Fields, and BoxTeacher \cite{cheng2022boxteacher} employing consistency regularization of the Mean-teacher technique to generate pseudo instance masks conditioned by ground-truth boxes. Although BS-2DIS is less challenging than \Problem, the structured and dense properties of 2D images that these regularization techniques imply do not hold in 3D point clouds, thus, we cannot trivially apply these methods in \Problem.

\myheading{Box-supervised 3D point cloud instance segmentation (\Problem)} aims to segment all instances of predefined classes, utilizing the supervision of axis-aligned 3D bounding boxes, which correspond to two 3D points per instance. Compared to point supervision techniques such as PointContrast \cite{xie2020pointcontrast} and CSC \cite{hou2021exploring}, \Problem~\cite{chibane2021box2mask,du2023weakly} is considered more appropriate in 3DIS segmentation with less supervision. This is because the former provides valuable information about object extent through its only one bounding box per instance whereas the latter relies on selecting specific labeled points, resulting in more sensitive results. Box2Mask \cite{chibane2021box2mask} was the first to introduce \Problem~utilizing point clustering to group points based on their predicted bounding boxes. 
WISGP \cite{du2023weakly} employs simple heuristics to propagate labels from determined points to undetermined points and uses the pseudo labels to train a fully-supervised 3DIS model. 
In contrast, our proposed approach utilizes uncertainty when predicting object masks with weak labels as additional pseudo labels. 
Furthermore, our approach incorporates Gaussian Processes to model pairwise similarity between regions, including determined-determined, determined-undetermined, and undetermined-undetermined relationships. This results in a more effective global label propagation than the local propagation between neighboring points utilized by \cite{du2023weakly}.

\myheading{Gaussian Process (GP) in 3D point cloud} methods including \cite{shin2017real, 5979818, 5152677} leverage GP to model the relationship among regions to predict semantic segmentation in the fully-supervised setting. On the other hand, our approach utilizes GP in the weakly-supervised setting of 3D instance segmentation, that is, to estimate the distribution of object masks from the provided GT 3D boxes to train a 3DIS network.

\section{Our Approach}
\label{sec:approach}

% \begin{figure*}[t]
%   \centering
%   \includegraphics[width=0.95\linewidth]{images/gp_arch.pdf}
%   % \vspace{-4pt}
%   \caption{Gaussian Processes. 
%   }
%   \label{fig:architecture}
%   % \vspace{-16pt}
% \end{figure*}

\myheading{Problem statement:} 
In training, we are given a 3D point cloud $\mathbf{P} \in \mathbb{R}^{N\times6}$ where $N$ is the number of points, and each point is represented by a 3D position and RGB color vector. We are also provided a set of 3D axis-aligned bounding boxes $\mathbf{B} \in \mathbb{R}^{K\times6}$ and their classes $\mathbf{L} \in \{1,\dotsc, C\}^{K \times 1}$, where $K$ is the number of instances and $C$ is the number of object classes, as the box-supervision. 
Each bounding box is represented by two corners with minimum and maximum XYZ coordinates.
Our approach, \Approach, attempts to generate pseudo object masks of these $K$ instances, $\mathbf{M} \in \{0,1\}^{K \times N}$, and use them to train a 3DIS network $\Phi$. 
In testing, given a new point cloud $\mathbf{P'} \in \mathbb{R}^{N'\times 6}$, $\Phi$ predicts the masks $\mathbf{\widehat{M}} \in \{0,1\}^{K' \times N'}$ of all $K'$ instances of the $C$ object classes.

The overall architecture of \Approach~is depicted in Fig.~\ref{fig:architecture}, which is a two-step approach that involves generating pseudo instance masks and their uncertainty values from box annotations with Gaussian Processes and training a 3DIS network with the resulting labels with a devised KL divergence loss along with the previous mask loss. 
% We can optionally iterative these two steps as a self-training strategy to further boost performance.

\subsection{Gaussian Processes as Pseudo Labelers}
\label{sec:gp_label}

\begin{figure}[t]
  \centering
  \includegraphics[width=.9\linewidth]{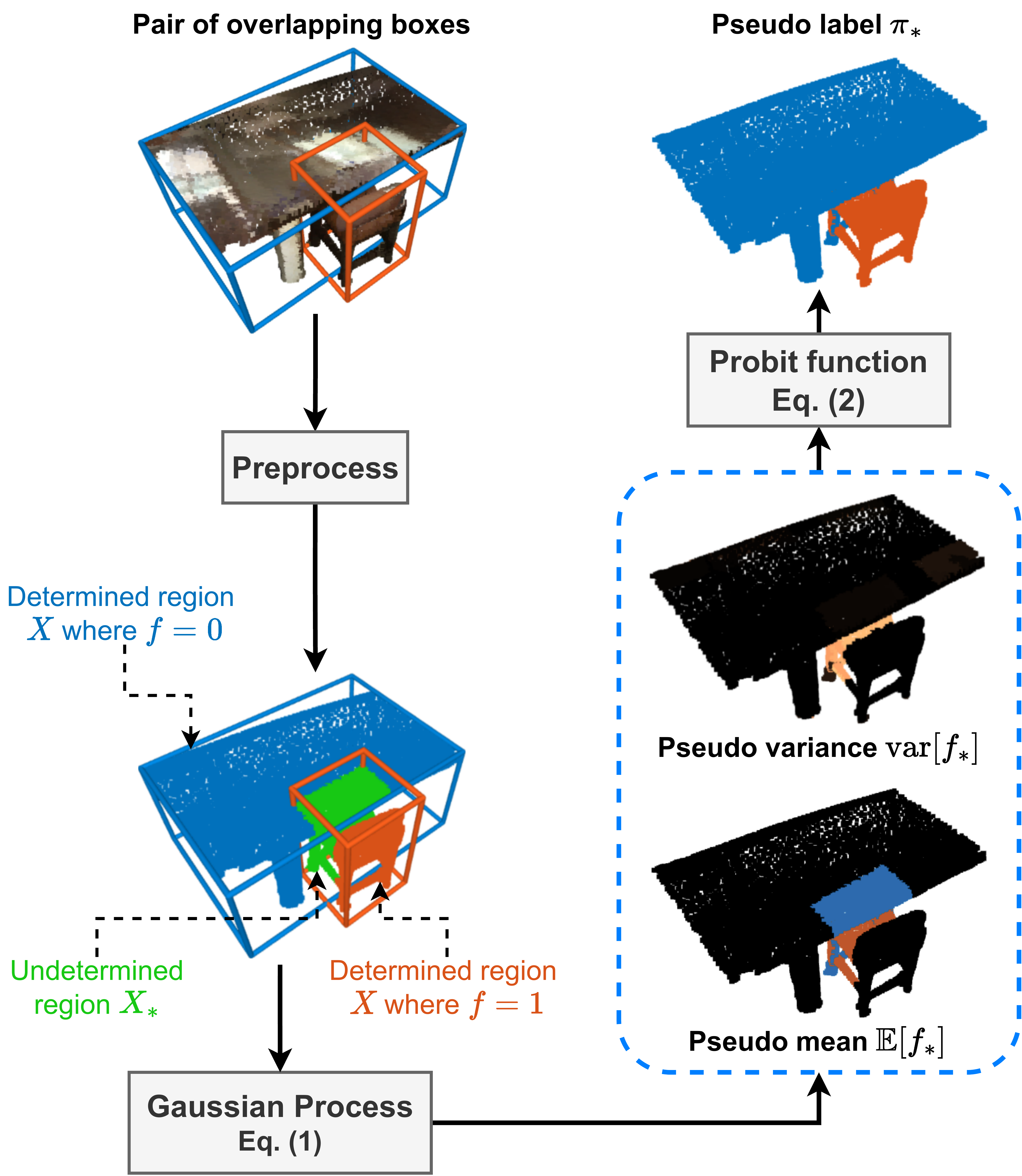}
   % \vspace{-4pt}
   \caption{\textbf{Our Gaussian Process}. For each pair of overlapping boxes, the determined and undetermined regions are identified and taken as input into a Gaussian Process to produce pseudo mean and variance values. Then the Probit function is utilized to output the posterior Bernoulli distribution as pseudo labels.
   }
   \label{fig:gp_arch}
   \vspace{-12pt}
\end{figure}

% We observe that due to the sparsity of 3D point clouds if a 3D point is within an axis-aligned bounding box representing an instance, the point probably belongs to this instance. Based on this geometric prior, we can roughly assign points to instances to generate pseudo object masks. However, as the axis-aligned bounding boxes do not tightly fit the specific shapes of complicated objects, there are many overlapped regions among these boxes, in which points can belong to multiple instances as visualized in Fig.~\ref{fig:teaser}. Thus, it is confusing when assigning points to instances. To address this issue, we propose using Gaussian Process (GP) as a probabilistic assigner to resolve the multi-box conflicts.
% We choose the Gaussian process as its advantages in our context are twofold. First, GP considers the complete relationships among regions so that the similarity between determined regions and the similarity between undetermined regions can affect the label propagation from determined to undetermined regions.
% Second, GP outputs a probabilistic distribution which allows uncertainty modeling of the pseudo labels.

% \son{We choose Gaussian processes as its advantages in our context are twofold. First, GP can be trained to assign pseudo labels with a small number of training samples, and thus avoid over-fitting. Second, GP outputs a probabilistic distribution which allows uncertainty modeling of the pseudo labels.}

We observed that 3D point clouds are sparse, and if a 3D point is within an axis-aligned bounding box representing an instance, it likely belongs to that instance. Using this geometric prior, we can roughly assign points to instances to generate pseudo object masks. However, as the axis-aligned bounding boxes do not accurately fit the complex shapes of objects, there are often overlapped regions among these boxes, leading to points belonging to multiple instances, as shown in Fig.~\ref{fig:teaser}. Consequently, assigning points to instances becomes challenging. To overcome this issue, we propose using Gaussian Process (GP) as a probabilistic assigner to resolve conflicts that arise from overlapping boxes.
We choose the Gaussian process for two reasons. Firstly, GP considers the complete relationships among regions, allowing the similarity between determined regions and the similarity between undetermined regions to affect label propagation from determined to undetermined regions. Secondly, GP outputs a probabilistic distribution, enabling the modeling of uncertainty in the pseudo labels.

% Unlike instance mask labels where each point belongs to at most one object, the axis-aligned box labels of different objects can overlap, resulting in the ambiguous point-object assignment. Thus, a simple heuristic rule such as assigning ambiguous points to the smallest box utilized in Box2Mask \cite{li2022box2mask} does not work as shown in Fig.~\ref{fig:teaser}. Therefore, in this paper, we propose an efficient Gaussian Process (GP) as a more effective pseudo labeler. 

% By careful analysis of the overlapping box labels in two 3DIS datasets, i.e. ScannetV2 \cite{dai2017scannet} and S3DIS \cite{armeni2017joint}, 9x\% of them are between two boxes, and the rest are among three or four boxes. Hence, we pay more attention to resolving the ambiguous point-object assignment between two boxes by employing a GPs on each pair of overlapping boxes. 

% There are two key reasons why GP is well-suited to our problem. First, it models the pairwise similarity among regions

 To begin, we divide the input point cloud into two non-overlapping sets: the \textit{determined set} and the \textit{undetermined set}. The determined set includes points that belong to at most one bounding box, and we assign these points to the corresponding label of the bounding box that encloses them. Points outside all bounding boxes are labeled as background. However, in the undetermined set, it is challenging to assign the correct labels to points that reside in the overlapped regions of bounding boxes. To solve this problem, we treat the assignment of points in the overlapped region of two boxes as a binary classification task and use the Gaussian Process as a probabilistic classifier.

While there are some regions that result from the intersections of more than two boxes, our analysis of the overlapping box labels in the ScanNetV2 \cite{dai2017scannet} and S3DIS \cite{armeni2017joint} 3DIS datasets shows that $95.4\%$ of cases involve only two boxes, and the remainder involve three or four boxes. In these infrequent cases, we select the pair with the largest overlap to use for the GP.
Additionally, both datasets include superpoints -- clusters of points grouped together based on their RGB color and position values. We can use these superpoints as elements in the GP rather than individual points, which can help reduce processing time, as utilized in Mask3D \cite{Schult23ICRA} and SPFormer \cite{sun2022superpoint}. Therefore, we will refer to both superpoints and individual points as regions going forward.

% \son{I revised this paragraph. Please read again.}
Our devised GP is illustrated in Fig.~\ref{fig:gp_arch}. Given two overlapping bounding boxes, the training data for GP is $n_1$ determined regions $\X \in \mathbb{R}^{n_1 \times 6}$ with their noise-free labels $\f \in \{0, 1\}^{n_1}$, or $p(\f) = \mathcal{N}(\f, \0)$.
The GP seeks to produce the outputs of $n_2$ testing undetermined regions $\X_* \in \mathbb{R}^{n_2 \times 6}$ including the underlying Gaussian distributions $p(\f_*) = \mathcal{N}(\mathbb{E}[\f_*], \text{var}[\f_*])$ of labels $\f_*$, and the pseudo labels $\boldsymbol{\pi_*}$ inferred from the distribution.  

In particular, we denote the output as the concatenation of the training labels $\mathbf{f}$ and the unknown $\mathbf{f}_*$, which follows the joint multivariate Gaussian distribution:
\begin{gather}
    \left(\begin{array}{c}
    \mathbf{f} \\
    \mathbf{f}_*
    \end{array}\right) \sim\mathcal{N}\left(\mathbf{0},\left(\begin{array}{cc}
    \K & \K_* \\
    \K_*^T & \K_{* *}
    \end{array}\right)\right),
\end{gather}
where $\K=\kappa(\X, \X) \in \mathbb{R}_+^{n_1\times n_1}, \K_*=\kappa(\X, \X_*) \in \mathbb{R}_+^{n_1\times n_2}, \K_{**}=\kappa(\X_*, \X_*) \in \mathbb{R}_+^{n_2\times n_2}$ are the covariance matrices that capture the relationship between determined regions, determined-undetermined regions, and undetermined regions, respectively. $\kappa\left(x, x^{\prime}\right)=s^2 \exp \left(-\frac{1}{2 \ell^2}\left(x-x^{\prime}\right)^2\right)$ is the radial basis kernel where $l$ and $s$ control the length scale and output scale.
We create separate a GP model for each pair of overlapping bounding boxes. The hyper-parameters, i.e., length scale $l$ and output scale $s$, are optimized by using the determined regions.

% \son{Maths to fix. Some symbols need more explanation: $\mathbf{f}$, $\mathbf{f}_*$, $f_*$. Define $\mathbf{X}$ and $\mathbf{X}_*$, eg $\mathbb{R}^{n \times 6}$ and $\mathbb{R}^{m \times 6}$ where $n+m$ is the total number of points in two bounding boxes. Define the integration bound in Equation 2. Maybe write the kernel function to show how s and l are used.}

The pseudo labels $\boldsymbol{\pi_*}$ can be computed as posterior:
\begin{gather}
    \boldsymbol{\pi_*}=p\left(\f_*=1 \mid \X_*, \X, \f \right) \approx \int \sigma\left(\f_*\right) p\left(\f_*\right) d \f_*, \nonumber \\ 
    \approx \sigma\left(\frac{\mathbb{E}\left[\f_*\right]}{\sqrt{1 + \frac{\pi}{8} \text{var}[\f_*])}} \right),
    % \label{Eq:probit}
\end{gather}
where the last approximation is the probit approximation, and $\sigma$ is sigmoid activation. 

For each object, the final binary mask $\m \in \{0, 1\}^{1 \times N}$ is obtained by attaching the regions $\X_*$ whose $\boldsymbol{\pi_*}\geq 0.5$ to the foreground regions of the object. Also, the mean map $\e \in [0, 1]^{1 \times N}$ is constructed by setting the mean of the determined regions to their labels and the mean of the undetermined regions to $\mathbb{E}[\f_*]$. Finally, the variance map $\v \in \mathbb{R}_+^{1 \times N}$ is constructed by setting the variance of the determined regions to $0$ and the variance of the undetermined regions to $\text{var}[\f_*]$.

% The Gaussian Process (GP) considers three types of pairwise relationships between regions: undetermined-undetermined regions captured by $\K_{**}$, determined-undetermined by $\K_*$, and determined-determined by $\K$, respectively, allowing for complete similarity modeling. This enables effective global label propagation from determined to undetermined regions. Additionally, the GP provides the mean $\mathbb{E}[f_*]$ and variance $\text{var}[f_*]$ of the underlying Gaussian. These statistics are valuable not only for correcting labels by human annotators (using variance as a measure of uncertainty), but also for supervising the 3DIS network along with the binary mask in the next step.

% A Gaussian Process is a generalization of the Gaussian probability distribution over random variables, where each random variable is considered as the function value f (x) at a specific input x \cite{RasmussenW06}. A GP is specified by a mean function m(x) and a covariance
% function k(x, x)
% ) as follows:

\subsection{Training a 3DIS Network with Pseudo Labels}
\label{sec:uncertain_loss}

After getting the pseudo masks $\M \in \{0, 1\}^{K \times N}$ from GP, we are ready to train any 3DIS network $\Phi$. However, to leverage the informative cues from the mean $\E \in [0, 1]^{K \times N}$ and variance $\V \in \mathbb{R}_+^{K \times N}$ maps  also inferred from GP, rather than predicting only instance masks $\widehat{\M}$, we can simply modify the last layer of the network to predict two additional outputs: the mean $\widehat{\E}$ and the variance $\widehat{\V}$ representing the predicted Gaussian distribution.  

For training the mask prediction $\widehat{\M}$, we use two loss functions: dice loss \cite{sudre2017generalised} and BCE loss following prior 3DIS work. For training the mean $\widehat{\E}$ and variance $\widehat{\V}$ predictions, we devise a new loss function based on KL divergence for each location $i$ as follows:
\begin{equation}
    L_{\text{KL}}(i) =
        \begin{cases}
			\log \frac{\hat{\v}_i}{\v_i}+\frac{\v_i^2+\left(\e_i-\hat{\e}_i\right)^2}{2 \hat{\v}_i^2}-\frac{1}{2}, & \text{if } \v_i > 0\\
            % 
            % (\e_i - \hat{\e}_i)^2 + (\v_i - \hat{\v}_i)^2, & \text{otherwise.}
            (\e_i - \hat{\e}_i)^2 + \hat{\v}_i^2, & \text{if } \v_i = 0,
	\end{cases}
\end{equation}
where $\e_i, \v_i$ are the mean and variance at location $i$. 
When the variance is positive, we want to match two Gaussian distributions using KL divergence. Otherwise, they are Dirac Delta functions, so the predicted mean is matched with the pseudo mean and the predicted variance is matched with the pseudo variance using the MSE loss. As will be shown in the experiments, using the $L_{\text{KL}}$ helps boost performance compared to only using mask loss.

\myheading{Self-training:} The feature for each point/superpoint can either be the input features (RGB color and position) or the pointwise deep feature extracted from a pretrained 3DIS network. Thus, after training the 3DIS network with the pseudo labels, we can utilize its pointwise deep features as $\X$ and $\X_*$, and then rerun the GP to obtain better pseudo labels. This strategy is referred to as \textit{self-training}.
% This process can be repeated $T$ times to further enhance the performance if needed.

\section{Experiments}
\label{sec:experiments}
% \subsection{Experimental Setup}
\myheading{Datasets.} We conduct experiments on two datasets: ScanNetV2 \cite{dai2017scannet} and S3DIS \cite{armeni2017joint}. \textit{ScanNetV2} consists of 1201, 312, and 100 scans with 18 object classes for training, validation, and testing, respectively. 
We report the evaluation results on the validation and test sets of ScanNetV2. 
The \textit{S3DIS} dataset contains 271 scenes from 6 areas with 13 categories. We use Area 1, 2, 3, 4, 6 for training and Area 5 for evaluation.
% We report evaluations for both Area 5 and 6-fold cross-validation as in the previous work.

\myheading{Evaluation metrics.} The average precision (AP) metrics commonly used in object detection and instance segmentation are adopted, including AP$_{50}$ and AP$_{25}$ are the scores with IoU thresholds of 50\% and 25\%, AP is the averaged score with IoU thresholds from 50\% to 95\% with a step size of 5\%, and
Box AP means the AP of the 3D axis-aligned bounding box prediction. Additionally, the S3DIS is also evaluated using mean coverage (mCov), mean weighed coverage (mWCov), mean precision (mPrec$_{50}$), and mean recall (mRec$_{50}$) with IoU threshold of 50\%.

\myheading{Implementation details.}
We implement our devised Gaussian Process by using GPytorch \cite{gardner2018gpytorch} to estimate $l, s$ and compute $\boldsymbol{\pi_*}, \mathbb{E}[\f_*], \text{var}[\f_*]$ efficiently. We leverage the Adam optimizer with a learning rate of 0.1. For reference, it takes approximately 5 hours to generate pseudo labels for the entire ScanNetV2 training set (1201 scenes) on a single V100.
We leverage our pseudo labels to train 5 different 3DIS methods, including PointGroup \cite{jiang2020pointgroup}, SSTNet \cite{liang2021instance}, SoftGroup \cite{vu2022softgroup}, ISBNet~\cite{ngo2023isbnet}, and SPFormer \cite{sun2022superpoint} based on their publicly released implementations. For methods that do not provide the code on S3DIS, we reproduce them based on the implementation details in their papers. All the models are trained from scratch and the hyper-parameters and the training details are kept the same as the original methods.

% We implement our model using PyTorch deep learning framework \cite{paszke2017automatic} and train it on 320 epochs with AdamW optimizer on a single V100 GPU. The batch size is set to 16. The learning rate is initialized to 0.004 and scheduled by a cosine annealing \cite{zhang2021point}. Following \cite{vu2022softgroup}, we set the voxel size to 0.02m for ScanNetV2 and S3DIS, and to 0.3m for STPLS3D due to its sparsity and much larger scale. In training, the scenes are randomly cropped at a maximum number of 250,000 points. In testing, the whole scenes are fed into the network without cropping. 
% We use the same backbone design as in \cite{vu2022softgroup}, which outputs a feature map of 32 channels. A stack of two layers of PA is used in the sampling-based instance-aware encoder. $\tau$ is set to 0.5. We set the ball query radius $r$ to 0.2 and 0.4 for these two layers and the number of neighbors $Q=32$ for both layers. 

\begin{table*}
\small
\setlength{\tabcolsep}{6pt}
\centering
% \begin{tabular}{lll>{\color{red}}c>{\color{red}}ccc>{\color{red}}c>{\color{red}}ccc}
\begin{tabular}{lllcc>{\color{gray!70}}c>{\color{gray!70}}ccc>{\color{gray!70}}c>{\color{gray!70}}c}
\toprule
\multirow{ 2}{*}{\textbf{Method}} & \multirow{ 2}{*}{\textbf{Sup.}} & \multirow{ 2}{*}{\textbf{Backbone}} & \multicolumn{4}{c}{\textbf{Test set}} & \multicolumn{4}{c}{\textbf{Val set}}\\
 & & & \textbf{AP} & \textbf{\% full} & \textbf{AP$_{50}$} & \textbf{AP$_{25}$} & \textbf{AP} & \textbf{\% full} & \textbf{AP$_{50}$} & \textbf{AP$_{25}$} \\ 
\midrule
% \rowcolor{gray}
% SGPN \cite{wang2018sgpn} & Masks & 4.9 & 14.3 & 26.1 \\
% GSPN\cite{yi2019gspn} & \\
% 3D-SIS \cite{hou2019sis} & CVPR 19 & 16.1 & 38.2 & 55.8 \\
% GSPN\cite{yi2019gspn} & \\
% MTML \cite{yang2019learning} & ICCV 19 & 28.2 & 54.9 & 73.1 \\
% 3D-BoNet \cite{yang2019learning} & Masks & 25.3 & 48.8 & 68.7 \\
% 3D-MPA \cite{engelmann20203d} & CVPR 20 & 35.5 & 61.1 & 73.7 \\

\rowcolor{gray!30} Mask3D \cite{Schult23ICRA} &  & Minkowski & 56.6 & - &	78.0 & 87.0 & 55.2 & - & 73.7 & 83.5  \\
% \rowcolor{gray!30} PointGroup \cite{jiang2020pointgroup} &  & SPConv  & 40.7 & 63.6 & 77.8 & 34.8 & 51.7 & 71.3 \\
% OccuSeg \cite{han2020occuseg} & CVPR 20 & 44.3 & 67.2 & 74.2 \\
% DyCo3D \cite{he2021dyco3d} & Masks & 39.5 & 64.1 & 76.1 \\
% PE \cite{zhang2021point} & CVPR 21 & 39.6 & 64.5 & 77.6 \\
% \rowcolor{gray!30} HAIS \cite{chen2021hierarchical} &  & SPConv & 45.7 & 69.9 & 80.3 & 43.5 & 64.4 & 75.6 \\ 
\rowcolor{gray!30} PointGroup \cite{jiang2020pointgroup} & & SPConv & 40.7 & - & 63.6 & 77.8 & 34.8 & - & 51.7 & 71.3 \\
\rowcolor{gray!30} SSTNet \cite{liang2021instance} &  & SPConv & 50.6 & - & 69.8 & 78.9 & 49.4 & - & 64.3 & 74.0 \\
\rowcolor{gray!30} SoftGroup \cite{vu2022softgroup} &  & SPConv & 50.4 & - & 76.1 & 86.5 & 46.0 & - & 67.6 & 78.9 \\
% RPGN \cite{dong2022rpgn} & ECCV 22 & 42.8 & 64.3 & 80.6 \\
% PointInst3D \cite{he2022pointinst3d} & Masks & 43.8 & - & - \\
% Di\&Co3D \cite{zhao2022divide} & ECCV 22 & 47.7 & 70.0 & 80.2 \\
% DKNet \cite{wu2022dknet} & Masks & 53.2 & 71.8 & 81.5 \\
\rowcolor{gray!30} ISBNet \cite{ngo2023isbnet} &  & SPConv &  55.9 & - & 76.3 & 84.5 & 54.5 & - & 73.1 & 82.5 \\
\rowcolor{gray!30} SPFormer \cite{sun2022superpoint} & \multirow{ -6}{*}{Mask} & SPConv & 54.9 & - & 77.0 & 85.1 & 56.3 & - & 73.9 & 82.9 \\
\midrule
% CSC \cite{hou2021exploring} & 20 points & 15.9 & 28.9 & 49.6 & 15.9 & 28.9 & 49.6 \\ 
% PointContrast \cite{xie2020pointcontrast} & 20 points & 12.4 & 25.9 & 47.4 & 12.4 & 25.9 & 47.4 \\ 
\rowcolor{gray!30} CSC \cite{hou2021exploring} &  & Minkowski & 29.3 & 51.8\% & 59.2 & 70.2 & 15.9 & 28.8\% & 28.9 & 49.6\\ 
\rowcolor{gray!30} PointContrast \cite{xie2020pointcontrast} & \multirow{ -2}{*}{Point} & Minkowski & 27.8 & 49.1\% & 47.1 & 64.5 & 27.8 & 50.4\% & 47.1 & 64.5 \\ 
\midrule
Box2Mask \cite{chibane2021box2mask} (stand-alone) & \multirow{ 3}{*}{Box} & Minkowski & 43.3 & - & 67.7 & 80.3 & 39.1 & - & 59.7 & 71.8 \\
% Box2Mask \cite{chibane2021box2mask} (stand-alone) &  & SPConv &  & - &  &  & 33.6 & - & 55.3 &	71.8  \\
% Box2Mask \cite{chibane2021box2mask} + PointGroup \cite{jiang2020pointgroup} &  & SPConv &  &  &  &  &  &  \\
% Box2Mask \cite{chibane2021box2mask} + SSTNet \cite{liang2021instance} &  & SPConv &  &  &  &  &  &  \\
WISGP \cite{du2023weakly} + PointGroup \cite{jiang2020pointgroup} &  & SPConv & - & - & - & - & 31.3 & 89.9\% & 50.2 & 64.9 \\
WISGP \cite{du2023weakly} + SSTNet \cite{liang2021instance}  &  & SPConv & - & - & - & - & 35.2 & 71.2\% & 56.9 & 70.2 \\
\midrule
\Approach~+ PointGroup \cite{jiang2020pointgroup} & \multirow{ 5}{*}{Box} & SPConv & 39.4 & 96.8\% & 62.3 & 74.5 & 33.4 & 96.0\% & 53.7 & 69.8 \\
\Approach~+ SSTNet \cite{liang2021instance} &  & SPConv & 45.8 & 90.5\% & 65.2 & 75.0 & 43.9 & 88.9\% & 60.1 & 70.8 \\
\Approach~+ SoftGroup \cite{vu2022softgroup} &  & SPConv & 42.1 & 83.5\% & 62.9 & 79.4 & 41.3	& 89.8\% & 62.7 & 77.3 \\
\Approach~+ ISBNet \cite{ngo2023isbnet} &  & SPConv & 49.3 & 88.2\% & 69.8 & 81.0 & 50.6 & 92.8\% & 69.1 & 79.3 \\
\Approach~+ SPFormer \cite{sun2022superpoint}  &  & SPConv & 48.2 & 87.7\% & 69.2  & 82.4 & 51.1 & 90.8\% & 70.4 & 79.9 \\
\bottomrule
\end{tabular}
% \vspace{-4pt}
\caption{\textbf{3D instance segmentation results on ScanNetV2 hidden test set and validation set in AP metrics.} For reference purposes, we show the results of methods that use other types of supervision, such as Mask or Point in \colorbox{gray!30}{gray}. The main metric for comparison is \textbf{AP}. The column \textbf{\% full} indicates the percentage of the current method's performance compared to its corresponding fully supervised counterpart in the \textbf{AP} column. For the backbone, Minkowski is much heavier than SPConv. For Point supervision, we used 200 points per scene (or 10-20 points per instance).}
% The columns in \textcolor{red}{red} are the main metrics for comparison. Column \textcolor{red}{\% full} shows the percentage of the current method compared with its corresponding fully supervised counterpart in column \textcolor{red}{AP}. For the backbone, Minkowski is much heavier than SPConv. For Point supervision, 200 points per scene (or 10-20 points per instance) are used.}
\label{tab:scannet_quanti}
% \vspace{-16pt}
\end{table*}

\begin{table}
\small
\setlength{\tabcolsep}{4pt}
\centering
% \begin{tabular}{ll>{\color{red}}c>{\color{red}}ccc}
\begin{tabular}{llcc>{\color{gray!70}}c>{\color{gray!70}}c}
\toprule
 \textbf{Method} & \textbf{Sup.} & \textbf{AP} & \textbf{AP$_{50}$} & \textbf{mPrec} & \textbf{mRec}  \\ 
\midrule
\rowcolor{gray!30} Mask3D \cite{Schult23ICRA} & & 56.6 & 68.4 & 68.7 & 66.3  \\
\rowcolor{gray!30} PointGroup \cite{jiang2020pointgroup} & & - & 57.8 & 61.9 & 62.1  \\
\rowcolor{gray!30} SSTNet \cite{liang2021instance} & & 42.7 & 59.3 & 65.6 & 64.2  \\
\rowcolor{gray!30} SoftGroup \cite{vu2022softgroup} &  & 51.6 & 66.1 & 73.6 & 66.6  \\
\rowcolor{gray!30} ISBNet \cite{ngo2023isbnet} & \multirow{-5}{*}{Mask} & 54.0 & 65.8 & 74.2 & 72.7 \\
% \rowcolor{gray!30} SPFormer \cite{sun2022superpoint} & \multirow{ -6}{*}{Mask} & - & 66.8 & 72.8 & 67.1 \\
\midrule
Box2Mask & \multirow{4}{*}{Box} & - & - & 66.7 & 65.5  \\
Box2Mask$^\ast$ &  & 43.6 & 54.6 & 64.4 & 67.4  \\
% Box2Mask \cite{chibane2021box2mask} (stand-alone) &  & SPConv &  & - &  &  & 33.6 & - & 55.3 \\
WISGP + PointGroup &  & 33.5 & 48.6 & 50.0 & 52.8  \\
WISGP + SSTNet  & & 37.2 & 51.0 & 44.3 & 56.7  \\
\midrule
\Approach~+ PointGroup & \multirow{4}{*}{Box} & 42.5 & 56.8 & 59.3 & 61.3 \\
\Approach~+ SSTNet & & 44.7 & 57.4 & 54.3 & 62.7 \\
% \Approach~+ SoftGroup & & 48.0 & 82.5 \\
\Approach~+ SoftGroup & & 47.0 & 62.1 & 64.8 & 67.0 \\
\Approach~+ ISBNet & & 50.5 & 61.2 & 66.7 & 72.4 \\
% \Approach~+ SPFormer & & 41.8 & 54.4 & &   \\
\bottomrule
\end{tabular}
% \vspace{-4pt}
\caption{\textbf{3DIS results on S3DIS on Area 5}. 
% The best results are in \textbf{bold} and the second best ones are in \underline{underlined}. 
The methods that use mask supervision are displayed in \colorbox{gray!30}{gray} and are solely for reference purposes. The primary metric for comparison is the \textbf{AP}. A $^\ast$ symbol indicates that we reproduced Box2Mask on the S3DIS dataset based on their public code. For the backbone of each method, please refer to Tab.~\ref{tab:scannet_quanti}.}
% The columns in \textcolor{red}{red} are the main metrics for comparison. $^\ast$ denotes that we reproduce Box2Mask on the S3DIS dataset based on their public code. For the backbone of each method, please refer to Tab.~\ref{tab:scannet_quanti}}
\label{tab:s3dis_quanti}
\vspace{-16pt}
\end{table}

\begin{figure*}[t]
  \centering
  \includegraphics[width=0.9\linewidth]{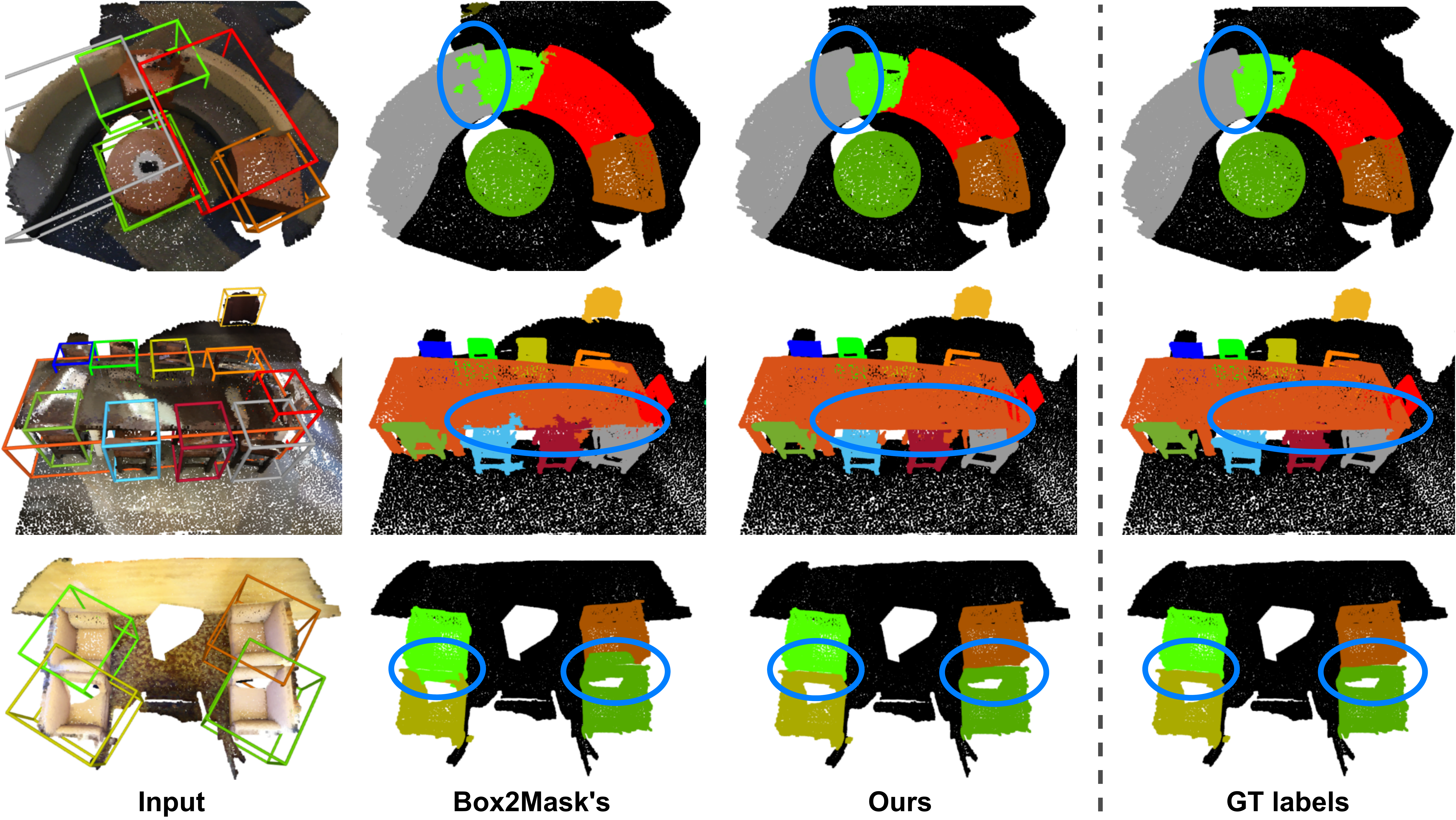}
% \vspace{-10pt}
   \caption{\textbf{Representative examples on ScanNetV2 training set}. Each row shows an example with the input and axis-aligned bounding box labels, Box2Mask \cite{chibane2021box2mask}'s pseudo labels, our pseudo labels, and GT labels, respectively. Our approach produces highly accurate instance masks, particularly in regions with overlapping GT bounding boxes (\textcolor{blue}{blue} circles).}
\vspace{-12pt}
   \label{fig:quali_scannetv2}
\end{figure*}

\subsection{Comparison to Prior Work}
Our direct comparison includes Box2Mask \cite{li2022box2mask} and WISGP \cite{du2023weakly}. Their details are specified in Sec.~\ref{sec:related_work}. 

\myheading{Quantitative results.} 
For ScanNetV2, we present the instance segmentation results for both the validation set and hidden test set in Tab.~\ref{tab:scannet_quanti}. It is obviously seen that our \Approach's versions of 3DIS methods outperform other box-supervised 3DIS methods by a significant margin on both sets, even with a smaller backbone (SPConv compared to Minkowski). Notably, our results are consistently comparable to SOTA fully supervised methods in AP, achieving about 90\%. These findings demonstrate the effectiveness of our approach and the potential of our pseudo labels for improving standard 3DIS methods.
For S3DIS, Tab.~\ref{tab:s3dis_quanti} presents the results on Area 5 of the S3DIS dataset. Our proposed \Approach~achieves superior performance compared to Box2Mask, with large margins in both AP and AP$_{50}$ when applied to SoftGroup and ISBNet. Additionally, when applied to PointGroup and SSTNet, our approach outperforms the WISGP's versions by a significant margin, demonstrating the robustness and effectiveness of our proposed pseudo labels.

\myheading{Qualitative results.}
We visualize the qualitative results of pseudo labels of Box2Mask \cite{chibane2021box2mask} and our method on ScanNetV2 training set in Fig.~\ref{fig:quali_scannetv2}. Our approach generates more precise pseudo instance masks than Box2Mask. Additionally, our method performs well even in challenging scenarios where objects are densely packed or share edges (2nd and 3rd row respectively), our method is able to accurately label points in overlapped regions.  

\subsection{Ablation Study}
We conduct ablation studies to justify the design choices of our proposed method. All these ablation experiments are conducted on ISBNet \cite{ngo2023isbnet} on the validation set of the ScanNetV2 dataset unless otherwise stated.

\myheading{Handling undetermined regions.} We first explore different techniques for handling undetermined regions (i.e., regions belonging to multiple boxes) in our proposed method. Tab.~\ref{tab:handle_undetermined} summarizes the results of our experiments.
In setting A, we evaluate the approach of ignoring undetermined regions during training and only using the determined regions as pseudo labels.
Next, inspired by the heuristics proposed by \cite{chibane2021box2mask}, we assign undetermined points to the smaller box. This approach, setting B, results in a +3.7 improvement in AP compared to ignoring undetermined regions. By replacing the previous heuristic rule with a simple linear classifier, setting C, we achieve 44.2 in AP.
In setting D1, we apply GP classification at the point level rather than the superpoint level. This approach significantly outperforms the heuristics-based approach from row 2 by a margin of +4 in AP.
Finally, in settings D2 and D3, we explore two variations of GP applied at the superpoint level. The D2 approach performs GP regression which predicts the mask value as a continuous value between 0 and 1, while the D3 approach performs GP classification directly on the superpoints. The latter achieves the highest results, with a +1 improvement in AP over the regression-based approach.

Furthermore, we evaluate the quality of pseudo masks by comparing them to GT labels in the \emph{training} set of ScanNetV2 using AP and AP$_{90}$ metrics. Tab.~\ref{tab:quanti_ps} shows that our GP-generated pseudo labels outperform setting A, B, and C. In E, we replace the labels of D3 predicted with high uncertainty by the GT labels so as to quantify the usefulness GP's uncertainty. This replacement leads to a notable improvement, 88.0 in AP while applying the same strategy for points with low uncertainty results in a lower AP of 86.3.

\myheading{Impact analysis of each component}
is summarized in Tab.~\ref{tab:ablation_components}.
In rows 1 and 2, we compare the performance with and without our GP-based pseudo labels. The results show a significant improvement in AP of up to +10 when our pseudo labels are used.
In row 3, we add a KL divergence loss during training with no additional cost to encourage the distribution of predicted masks to match the distribution of pseudo labels. This brings a further improvement of +0.3 in AP.
In row 4, we incorporate self-training to refine the quality of our pseudo labels, resulting in a higher quality of training data and a performance boost of +0.8 in AP.
Finally, in row 5, we combine all the components to produce our proposed approach, which achieves the best performance.

\begin{table}
\small
\setlength{\tabcolsep}{4.5pt}
\centering
\begin{tabular}{llcc}
\toprule
 & \textbf{Handling of undetermined points} & \textbf{AP} & \textbf{AP$_{50}$}  \\
\midrule
 & A: No pseudo labels in overlapped regions & 38.1 & 59.1 \\
 & B: Box2Mask: assign points to smaller boxes  & 41.8 & 64.8 \\
 &  C: Linear Classifier with points & 44.2 & 64.5 \\
\midrule
\multirow{3}{*}{\rotatebox{90}{\textbf{GaPro}}} & D1: GP Classification with points &  45.7 &	67.2 \\
 & D2: GP Regression with superpoints &  47.8 & 67.7 \\
 & D3: \textbf{GP Classification with superpoints} & \textbf{48.9} & \textbf{68.4} \\
\bottomrule
\end{tabular}
\vspace{-4pt}
\caption{Handling the undetermined regions to produce pseudo labels.}
\label{tab:handle_undetermined}
\vspace{-8pt}
\end{table}

\begin{table}
    \small
    \setlength{\tabcolsep}{4pt}
    \centering
    \begin{tabular}{lcc}
    \toprule
    \textbf{Handling of undetermined points} & \textbf{AP}  & \textbf{AP$_{90}$} \\
    \midrule
    A: No pseudo labels in overlapped regions    & 53.6 & 22.5   \\ 
    B: Box2Mask: assign points to smaller box & 64.4 & 27.6  \\
    C: Linear classifier with points & 69.4 & 34.1 \\
    \midrule
    D3: \textbf{\Approach~(ours)} & \textbf{85.9} & \textbf{63.1}\\
    E: Ours w/ uncertainty-guided GT replacement & 88.0 & 67.2 \\
    \bottomrule
    \end{tabular}      
    \vspace{-6pt}
    \caption{Quality of pseudo labels. We compute APs on the GT labels in the training set of ScanNetV2.} 
    \vspace{-8pt}
    \label{tab:quanti_ps}
\end{table}

\begin{table}
\small
\setlength{\tabcolsep}{3pt}
\centering
\begin{tabular}{cccccc}
\toprule
% \textbf{Handle} & \multirow{ 2}{*}{\textbf{KL Loss}} & \multirow{ 2}{*}{\textbf{Self-train.}} & \multirow{ 2}{*}{\textbf{AP}} & \multirow{ 2}{*}{\textbf{AP$_{50}$}} & \multirow{ 2}{*}{\textbf{AP$_{25}$}} \\
% \textbf{undetermined} \\
\textbf{Our pseudo labels} & \textbf{KL loss} & \textbf{Self-train.} & \textbf{AP} & \textbf{AP$_{50}$} & \textbf{AP$_{25}$} \\
\midrule
% \multirow{2}{*}{NMC} & & & & 33.6 & 55.3 & 71.8 \\
%  & \checkmark & & & 40.2 & 60.1 & 73.7 \\
% \midrule
 & & & 38.1 & 59.1 & 72.7 \\
\checkmark &  & & 48.9 & 68.4 & 79.0 \\
\checkmark & \checkmark & & 49.2 & 68.1 & 78.5 \\
\checkmark &  & \checkmark & 50.0 & 68.3 & 79.0 \\
\midrule
\checkmark & \checkmark  & \checkmark & \textbf{50.6} & \textbf{69.1} & \textbf{79.3} \\ 
\bottomrule
\end{tabular}
\vspace{-4pt}
\caption{Impact of our \Approach's components.
% \textbf{NMC}: Using Non-maximum-clustering \cite{chibane2021box2mask},
% \textbf{E2E}: Using end-to-end method \cite{ngo2023isbnet},
\textbf{Our Pseudo Labels}: the proposed pseudo labels in Sec.~\ref{sec:gp_label},
\textbf{KL Loss}: KL divergence loss,
\textbf{Self-train.}: Self-training.}
\label{tab:ablation_components}
\vspace{-10pt}
\end{table}

% \begin{table}
% \small
% \setlength{\tabcolsep}{4.5pt}
% \centering
% \begin{tabular}{cccc}
% \toprule
% \textbf{Handle Undetermined Set} & \textbf{AP} & \textbf{AP$_{50}$} & \textbf{AP$_{25}$} \\
% \midrule
% % \multirow{2}{*}{NMC} & & & & 33.6 & 55.3 & 71.8 \\
% %  & \checkmark & & & 40.2 & 60.1 & 73.7 \\
% % \midrule
% Ignore (no pseudo labels)  & 38.1 & 59.1 & 72.7 \\
% Assign points to smaller box  & 41.8 & 64.8 & 79.1 \\
% GP Classification with points &  45.7 &	67.2 &	78.9 \\
% GP Classification with superpoints &  47.8 & 67.7 & 79.0 \\
% GP Regression with superpoints &  48.9 & 68.4 & 79.0 \\
% \bottomrule
% \end{tabular}
% \vspace{-4pt}
% \caption{Different ways to handle the undetermined set to produce pseudo labels.}
% \label{tab:handle_undetermined}
% \vspace{-4pt}
% \end{table}

% \myheading{Performance of our GP on Box2Mask}. To further prove the robustness of our proposed method, we replace the heuristic rule in Box2Mask with our pseudo labels. 

\myheading{Study on the configuration of GP} is represented in Tab.~\ref{tab:ablation_config_gp}. We found that allowing the GP parameters, i.e., length scale $l$ and output scale $s$, to be learned resulted in a performance gain of more than 2 in AP. Furthermore, running GP on the superpoint level led to an additional improvement of 2 in AP compared to the version with point level. 

\begin{table}[t]
    \small
    \setlength{\tabcolsep}{4pt}
    \centering
    \begin{tabular}{cccc}
    \toprule
    \textbf{GP parameters} & \textbf{Superpoint} & \textbf{AP} & \textbf{AP$_{50}$} \\ 
    \midrule
    Fixed &  & 46.3 & 66.3 \\
    Fixed & \checkmark & 48.0 & 67.2 \\
    Learnable &  & 48.5 & 67.7 \\
    % \midrule
    Learnable & \checkmark & \textbf{50.6} & \textbf{69.1} \\
    \bottomrule
    \end{tabular}
    % \vspace{-4pt}
    \caption{Different configurations of GP. For fixed parameters, we set the length $l=0.5$ and output scales $s=1$.}
    \label{tab:ablation_config_gp}
    \vspace{-10pt}
\end{table}

\myheading{Study on the features of GP} is shown in Tab.~\ref{tab:ablation_feat_gp}. The first two rows present the results when we use only the position and normal of the point cloud as input to GP. When using \textit{deep} features obtained from a 3DIS network pretrained on our pseudo labels, the performance improved by +1.6 in AP.

\myheading{Study on different losses to use with uncertainty values} is reported in Tab.~\ref{tab:ablation_klloss}. In row 2, simply using MSE loss for all points brings no difference to the overall performance. Our KL divergence loss helps improve the AP by 0.6 in row 3.

% \begin{table}[t]
%     \small
%     \setlength{\tabcolsep}{7pt}
%     \centering
%     \begin{tabular}{lcc}
%     \toprule
%     \textbf{Feature Type} & \textbf{AP}     & \textbf{AP$_{50}$} \\ 
%     \midrule
%     Position & 49.0 & 68.3\\ 
%     Position + Normal & 49.5 & 68.5 \\
%     Deep & \textbf{51.1} & \textbf{69.5} \\
%     % \checkmark & \checkmark & \checkmark & 50.5 & 68.9 \\
%     \bottomrule
%     \end{tabular}
%     % \vspace{-4pt}
%     \caption{The impact of different types of features.}
%     \label{tab:ablation_feat_gp}
% \end{table}

% \myheading{The impact of the KL divergence loss}

% \begin{table}[t]
%     \small
%     \setlength{\tabcolsep}{7pt}
%     \centering
%     \begin{tabular}{lccc}
%     \toprule
%      & \textbf{AP}     & \textbf{AP$_{50}$} & \textbf{AP$_{25}$} \\ 
%     \midrule
%     Soft label & 50.1 & 68.5 & 78.2 \\
%     MSE & & & \\ 
%     \midrule
%     KL Loss &  &  \\
%     \bottomrule
%     \end{tabular}
%     % \vspace{-4pt}
%     \caption{Performance of different types of loss functions for the variance of GP labels.}
%     \label{tab:pseudo_label_type}
% \end{table}

\begin{table}[t]
    \parbox{.48\linewidth}{
        \small
        \setlength{\tabcolsep}{4.5pt}
        \centering
        \begin{tabular}{lcc}
        \toprule
        \textbf{Feature type} & \textbf{AP}     & \textbf{AP$_{50}$} \\ 
        \midrule
        % Pos. & 49.0 & 68.3\\ 
        % Pos. + Norm. & 49.5 & 68.5 \\
        % Deep & \textbf{51.1} & \textbf{69.5} \\
        Pos. & 48.5 & 67.9 \\ 
        Pos. + Norm. & 49.0 & 68.1 \\
        Deep & \textbf{50.6} & \textbf{69.1} \\
        % \checkmark & \checkmark & \checkmark & 50.5 & 68.9 \\
        \bottomrule
        \end{tabular}
        % \vspace{-4pt}
        \caption{Impact of different features to GP.}
        \label{tab:ablation_feat_gp}
    }
    \hfill
    \parbox{.47\linewidth}{
        \small
        \setlength{\tabcolsep}{6.2pt}
        \centering
        \begin{tabular}{lcc}
        \toprule
        \textbf{Loss type} & \textbf{AP}     & \textbf{AP$_{50}$} \\ 
        \midrule
        None & 50.0 & 68.3 \\
        MSE & 49.9 & 68.5 \\ 
        KL Loss & \textbf{50.6} & \textbf{69.1} \\
        \bottomrule
        \end{tabular}
        % \vspace{-4pt}
        \caption{Different losses to use with uncertainty values.}
        \label{tab:ablation_klloss}
        }
\end{table}

\myheading{3D Object Detection Results.} Our approach infers axis-aligned 3D bounding boxes, i.e., by taking the min and max coordinates of each dimension of the predicted instance masks, and we compare our results with other 3D object detection methods in Tab.~\ref{tab:3dod_results}. Notably, our findings demonstrate that when trained with the same level of annotations, the \Approach~versions of 3DIS methods can outperform SOTA 3D object detection methods by a significant margin, achieving a Box AP$_{50}$ increase of +8.6. 

\begin{table}
\small
\setlength{\tabcolsep}{5pt}
\centering
\begin{tabular}{lccc} %{l>{\color{red}}c>{\color{red}}c}
\toprule
\textbf{Method} & \textbf{Venue} & \textbf{Box AP$_{50}$}     & \textbf{Box AP$_{25}$} \\ 
\midrule
VoteNet \cite{qi2019deep} & ICCV 19 & 33.5 & 58.6 \\
3DETR \cite{misra2021end} & ICCV 21 & 47.0 & 65.0 \\
GroupFree \cite{liu2021group} & ICCV 21 & 52.8 & 69.1 \\
RGBNet \cite{wang2022rbgnet} & CVPR 22 & 55.2 & 70.6 \\
HyperDet3D \cite{zheng2022hyperdet3d} & CVPR 22 & 57.2 & 70.9 \\
FCAF3D \cite{rukhovich2022fcaf3d} & ECCV 22 & 57.3 & 71.5 \\
% CAGroup3D \cite{wang2022cagroup3d} & 61.3 & 75.1 \\
\midrule
\Approach~+ PointGroup & -  & 52.6 & 66.0 \\
\Approach~+ SSTNet & -  & 57.8 & 67.8 \\
\Approach~+ SoftGroup & - & 60.2 & 73.4 \\
\Approach~+ SPFormer & - & 65.9 & \textbf{78.9} \\
\Approach~+ ISBNet & - & \textbf{67.0} & 77.1 \\
\bottomrule
\end{tabular}
\vspace{-4pt}
\caption{3D object detection results on ScanNetV2 val set.}
\label{tab:3dod_results}
\vspace{-14pt}
\end{table}

\section{Discussion}
\label{sec:conclusion}
\begin{figure}[t]
  \centering
  \includegraphics[width=0.9\linewidth]{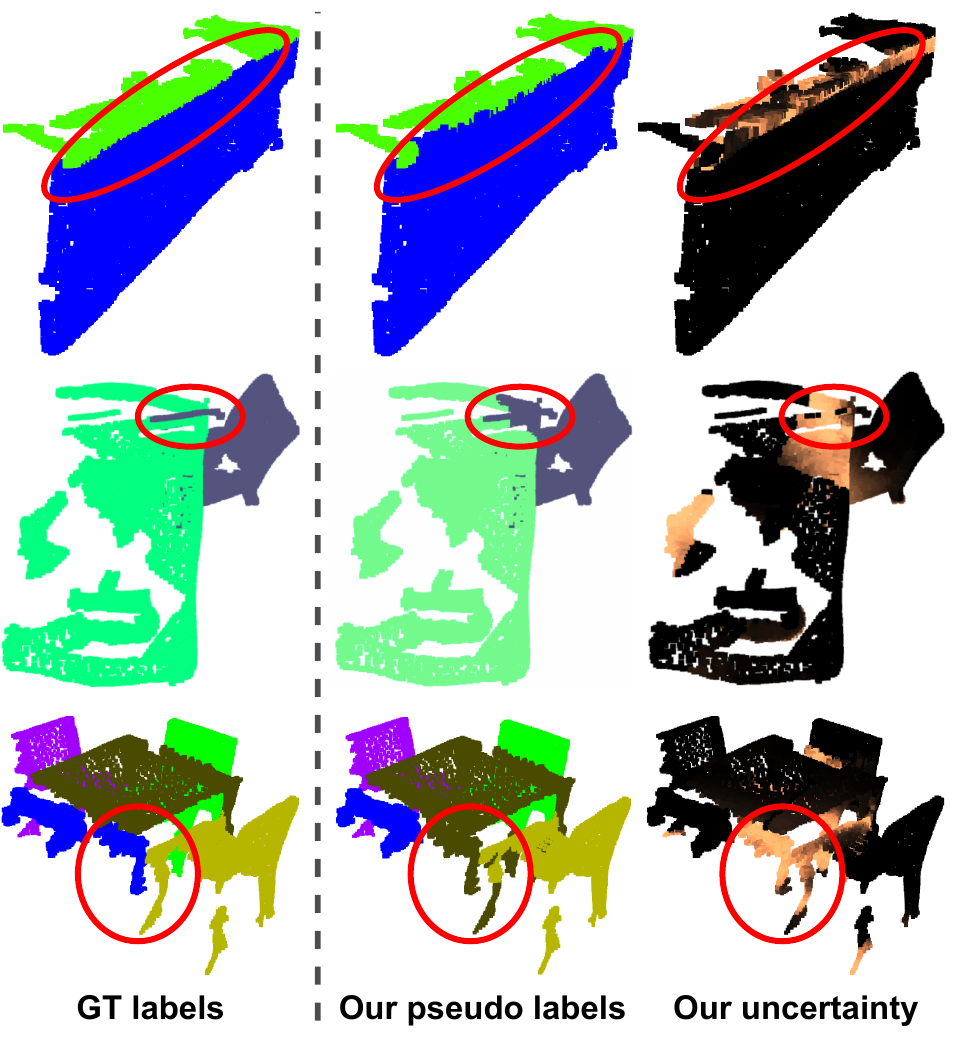}
   % \vspace{-4pt}
   \caption{Examples of our imperfect GP pseudo labels with their informative uncertainty values for annotators to correct.
   }
   \label{fig:failure_cases}
   \vspace{-12pt}
\end{figure}

\myheading{Limitations:}
% \myheading{Performance when using noisy boxes}. 
% so that each point belongs to a single box if it is not in the overlapping regions of the bounding boxes. 
Although our approach assumes accurately annotated bounding boxes for all considered objects to generate pseudo labels, this assumption is no longer valid when the bounding boxes are noisy or incomplete. To simulate such scenarios, we conducted two experiments: (1) adding Gaussian noise to the coordinates of two defining corners of GT boxes to create noisy bounding boxes, and (2) randomly dropping accurate GT boxes to create incomplete GT bounding boxes (Tabs. \ref{tab:ablation_corner} and \ref{tab:ablation_drop}, respectively). As shown in our experiments, the quality of the bounding boxes can significantly affect the accuracy of our pseudo labels. Moreover, even with accurate GT boxes, our pseudo labels may not be perfect in cases where there are overlapping boxes between adjacent objects or connecting objects with ambiguous shapes, as exemplified in Fig.~\ref{fig:failure_cases}. In such cases, our uncertainty values can provide useful indications for annotators to correct the pseudo labels.

\begin{table}[t]
    \small
    \setlength{\tabcolsep}{5pt}
    \centering
    \begin{subtable}{0.49\linewidth}
      \centering
        \begin{tabular}{ccc}
        \toprule
        \textbf{Cor. noise} & \textbf{AP} & \textbf{AP$_{50}$} \\ 
        \midrule
        2cm & 48.3 & 67.4\\
        5cm & 45.0 & 65.7 \\ 
        10cm & 43.0 & 64.2 \\ 
        10\% dim & 34.3 & 58.6 \\ 
        20\% dim & 21.0 & 43.5 \\ 
        \bottomrule
        \end{tabular}
      \caption{GT boxes with corner noises.}
      \label{tab:ablation_corner}
    %   \vspace{5}
    \end{subtable}
    % \vspace{20pt}
    \begin{subtable}{0.49\linewidth}
      \centering
        \begin{tabular}{ccc}
        \toprule
        \textbf{Drop rate} & \textbf{AP} & \textbf{AP$_{50}$} \\ 
        \midrule
        % 2\% & x & x \\
        5\% & 49.6 & 68.2 \\ 
        10\% & 49.1 & 68.1 \\ 
        20\% & 48.2 & 66.7 \\ 
        50\% & 41.6 & 61.2 \\
        80\% & 30.6 & 48.6 \\
        \bottomrule
        \end{tabular}
      \caption{Dropping GT boxes.}
      \label{tab:ablation_drop}
    \end{subtable}
    % \vspace{-12pt}
    \caption{Results drop with noisy and incomplete boxes.}
    \label{tab:ablation_noise}
\end{table}

% \myheading{Different types of inputs (points/superpoints)}

% \myheading{Manually Correct labels with uncertainty}

% \myheading{Failure cases of our GP pseudo labelers} are visualized in Fig.~\ref{fig:failure_cases}. 
% They mainly occur (1) at the common edge of two adjacent instances (row 1) and (2) at connecting objects with ambiguous shapes (rows 2 and 3). In these cases, the uncertainty (GP variance) provides useful information that can guide annotators in correcting the pseudo labels.

\myheading{Conclusion:}
In this work, we have introduced \Approach, a novel approach for instance segmentation on 3D point clouds using axis-aligned 3D bounding box supervision. Our approach generates high-quality pseudo instance masks along with associated uncertainty values, leading to superior performance compared to previous weakly supervised methods and competitive performance with SOTA fully supervised methods, achieving an accuracy of approximately 90\%. Additionally, our method's robustness has allowed for the easy adaptation of various fully supervised to weakly supervised versions using our pseudo labels, showing its potential for applications where obtaining fine-grain labels is costly.

{\small
\bibliographystyle{ieee_fullname}
\bibliography{egbib}
}

\newpage

\section{Supplementary Material}
In this supplementary material, we provide:
\begin{itemize}[noitemsep,topsep=0pt]
    \setlength\itemsep{0em}
    \item Analysis on the impact of using superpoints in Gaussian Processes (\cref{sec:gp_analysis}).
    \item Results on S3DIS with 6-fold cross validation (\cref{sec:s3dis_6fold}).
    \item Runtime statistics including model parameters and training time (\cref{sec:param_analysis}). 
    \item Per-class AP on the ScanNetV2 validation set and hidden test set (\cref{sec:perclass_scannet}).
    \item More qualitative results of our approach on all test datasets (\cref{sec:more_qualitative}).
\end{itemize}

\subsection{Impact of Superpoints in Gaussian Processes}
\label{sec:gp_analysis} 

Due to the typically large number of points per instance in 3D point clouds, running Gaussian Process (GP) directly on a point level is often impractical. For example, the ScanNetV2 \cite{dai2017scannet} dataset has around 1K-10K points per instance. To address this issue, we developed a point-level version of GP (row 3 in Tab. 3 in the main paper), which subsamples the top 800 nearest points from the determined points for each undetermined region. This version requires approximately 15 hours to generate pseudo labels for the entire ScanNetV2 training set on a single V100 GPU.

On the contrary, for the superpoint level version, the number of superpoints typically ranges from 10 to 1K, which allows us to run GP directly on all the superpoints. This version generates the pseudo labels in just 5 hours, as reported in the main paper.
We found that the accuracy of our method when using superpoints is improved compared to using points because we can consider all superpoints and the superpoints are well aligned to the instance boundaries. These results are shown in \cref{tab:detail_spp_gp}.

\subsection{Quantitative Results on S3DIS 6-fold Cross validation}
\label{sec:s3dis_6fold}

\cref{tab:quanti_s3dis_6fold} summarizes the results on 6-fold cross-validation of the S3DIS \cite{armeni2017joint} dataset. We observe the same trend as the results on Area 5 in Tab. 2 of the main paper.

\subsection{Run-time Statistics} 
\label{sec:param_analysis}
\cref{tab:param_detail} shows the parameters of the models and the training time of multiple methods on the ScanNetV2 dataset. For the \textbf{training time}, all the models are trained with batch size=8 on a single V100 GPU without mixed-precision training (FP16=False), and the other training details are kept the same as the original models.
Our method generates pseudo-labels and can be plugged and played with any instance segmentation method. Therefore, there is no run-time overhead when applying our pseudo-labels for training compared to the full supervision case. The training time difference in \cref{tab:param_detail} is mainly due to running time variations in data loaders and network optimizations.

\begin{table}[ht]
\small
\setlength{\tabcolsep}{4.5pt}
\centering
\begin{tabular}{lccc}
\toprule
 & \textbf{AP} & \textbf{AP$_{50}$} & \textbf{Gen. Time} \\
\midrule
GP with points &  45.7 &	67.2 & 15 hours \\
\textbf{GP with superpoints} & \textbf{48.9} & \textbf{68.4} & \textbf{5 hours}\\
\bottomrule
\end{tabular}
\vspace{-4pt}
\caption{Handling the undetermined regions to produce pseudo labels.}
\label{tab:detail_spp_gp}
\vspace{-10pt}
\end{table}

\begin{table}
\small
\setlength{\tabcolsep}{8pt}
\centering
\begin{tabular}{lccc}
\toprule
\textbf{Method} & \textbf{Sup.} & \textbf{AP} & \textbf{AP$_{50}$} \\ 
\midrule
\rowcolor{gray!30} PointGroup \cite{jiang2020pointgroup} &  & - & 64.0  \\
\rowcolor{gray!30} SoftGroup \cite{vu2022softgroup} &  & 54.4 & 68.9 \\
\rowcolor{gray!30} ISBNet \cite{ngo2023isbnet} & \multirow{-3}{*}{Mask} & 60.8 & 70.5 \\
% \midrule
% Box2Mask & Box & x & x   \\
\midrule
\Approach~+ PointGroup & \multirow{3}{*}{Box} & 46.0  & 60.4  \\
\Approach~+ SoftGroup &  & 51.4 & 65.8 \\
\Approach~+ ISBNet &  & 51.5 & 66.8  \\
\bottomrule
\end{tabular}
% \vspace{-4pt}
\caption{Results on S3DIS with 6-fold cross validation.}
\label{tab:quanti_s3dis_6fold}
\end{table}

\begin{table}
\small
\setlength{\tabcolsep}{3.5pt}
\centering
\begin{tabular}{lcccc}
\toprule
\multirow{ 2}{*}{\textbf{Method}} & \multirow{ 2}{*}{\textbf{Sup.}} & \textbf{\# of} & \textbf{Training} & \multirow{ 2}{*}{\textbf{AP}}  \\ 
 & & \textbf{params} & \textbf{time} & \\
\midrule
\rowcolor{gray!30} PointGroup \cite{jiang2020pointgroup} & & 7.7M & 32H & 34.8   \\
\rowcolor{gray!30} SSTNet \cite{liang2021instance} & & 113.2M & 57H & 49.4  \\
\rowcolor{gray!30} SoftGroup \cite{vu2022softgroup} & & 30.8M & 47H & 46.0 \\
\rowcolor{gray!30} ISBNet \cite{ngo2023isbnet} & & 31.1M & 39H & 54.5 \\
\rowcolor{gray!30} SPFormer \cite{sun2022superpoint} &  \multirow{-5}{*}{Mask} & 17.6M & 51H & 56.3 \\
\midrule
Box2Mask \cite{chibane2021box2mask} &  \multirow{3}{*}{Box} & 37M & 101H & 39.1   \\
WISGP \cite{du2023weakly} + PointGroup & & -  & - & 31.3   \\
WISGP \cite{du2023weakly} + SSTNet & & - & - & 35.2  \\
\midrule
\Approach~+ PointGroup & \multirow{5}{*}{Box} & 7.7M & 32H  & 33.4  \\
\Approach~+ SSTNet & & 113.2M & 57H & 43.9 \\
\Approach~+ SoftGroup & & 30.8M & 48H & 41.3 \\
\Approach~+ ISBNet & & 31.1M & 40H & 50.6  \\
\Approach~+ SPFormer & & 17.6M & 52H & 51.1   \\
\bottomrule
\end{tabular}
% \vspace{-4pt}
\caption{Models' parameters and training time on the ScanNetv2 validation set.}
\vspace{-10pt}
\label{tab:param_detail}
\end{table}

\subsection{Per-class AP on the ScanNetV2 dataset} 
\label{sec:perclass_scannet}

We report the detailed results of the 18 classes on the ScanNetV2 validation set and hidden test set in \cref{tab:scannet_val_detail} and \cref{tab:scannet_test_detail}, respectively.

\begin{table*}[t]
\small
\setlength{\tabcolsep}{2.6pt}
\centering
\begin{tabular}{l|c|ccccccccccccccccccccc}
\toprule
Method & AP & \rotatebox{90}{bathtub} & \rotatebox{90}{bed} & \rotatebox{90}{bookshe.}      & \rotatebox{90}{cabinet}  & \rotatebox{90}{chair} & \rotatebox{90}{counter}  & \rotatebox{90}{curtain}  & \rotatebox{90}{desk}  & \rotatebox{90}{door}  & \rotatebox{90}{other}  & \rotatebox{90}{picture}  & \rotatebox{90}{fridge} & \rotatebox{90}{s.curtain}  & \rotatebox{90}{sink}  & \rotatebox{90}{sofa} & \rotatebox{90}{table} & \rotatebox{90}{toilet} & \rotatebox{90}{window} \\ 
\midrule
\rowcolor{gray!30} PointGroup \cite{jiang2020pointgroup}  & 34.8 & 59.7 & 37.6 & 26.7 & 25.3 & 71.2 & 6.9 & 26.6 & 14.0 & 22.9 & 33.9 & 20.8 & 24.6 & 41.6 & 29.8 & 43.4 & 38.5 & 75.8 & 27.5  \\
\rowcolor{gray!30} SSTNet \cite{liang2021instance}  & 49.4 & 77.7 & 56.6 & 25.8 & 40.6 & 81.8 & 22.5 & 38.4 & 28.1 & 42.9 & 52.0 & 40.3 & 43.8 & 48.9 & 54.9 & 52.6 & 55.7 & 92.9 & 34.3 \\
\rowcolor{gray!30} SoftGroup \cite{vu2022softgroup} & 46.0 & 66.8 & 48.6 & 32.6 & 37.9 & 72.6 & 14.5 & 37.8 & 27.8 & 35.4 & 42.2 & 34.3 & 56.4 & 57.6 & 39.8 & 47.8 & 54.3 & 88.7 & 33.2 \\
\rowcolor{gray!30} ISBNet \cite{ngo2023isbnet} & 54.5 & 76.3 & 58.0 & 39.3 & 47.7 & 83.1 & 28.8 & 41.8 & 35.9 & 49.9 & 53.7 & 48.6 & 51.6 & 66.2 & 56.8 & 50.7 & 60.3 & 90.7 & 41.1  \\ 
\rowcolor{gray!30} SPFormer \cite{sun2022superpoint} & 56.3 & 83.7 & 53.6 & 31.9 & 45.0 & 80.7 & 38.4 & 49.7 & 41.8 & 52.7 & 55.6 & 55.0 & 57.5 & 56.4 & 59.7 & 51.1 & 62.8 & 95.5 & 41.1 \\
\midrule
Box2Mask \cite{chibane2021box2mask} & 39.5 & 70.6 & 41.7 & 23.1 & 27.4 & 73.8 & 8.8 & 31.0 & 14.4 & 27.1 & 45.1 & 31.5 & 34.3 & 44.3 & 46.0 & 51.1 & 31.4 & 83.6 & 25.9 \\
WISGP \cite{du2023weakly} + PointGroup & 31.3 & 40.2 & 34.7 & 26.2 & 27.2 & 69.1 & 5.9 & 19.9 & 8.7 & 18.2 & 30.9 & 26.2 & 30.7 & 33.1 & 23.8 & 33.9 & 39.1 & 73.7 & 22.4 \\
WISGP \cite{du2023weakly} + SSTNet & 35.2 & 45.5 & 32.8 & 23.8 & 30.4 & 75.3 & 8.8 & 23.9 & 17.6 & 27.8 & 33.0 & 28.4 & 31.4 & 23.1 & 32.9 & 42.7 & 39.4 & 83.4 & 25.9 \\
\midrule
\Approach~+ PointGroup & 33.4 & 46.8 & 58.1 & 32.4 & 31.4 & 63.1 & 21.8 & 26.5 & 36.2 & 20.3 & 27.4 & 20.6 & 25.8 & 20.9 & 18.5 & 48.2 & 41.6 & 65.0 & 18.4 \\
\Approach~+ SSTNet & 43.9 & 70.2 & 67.0 & 19.0 & 38.8 & 75.4 & 21.3 & 36.2 & 44.1 & 37.8 & 45.9 & 34.5 & 35.6 & 32.0 & 44.8 & 53.0 & 54.3 & 76.0 & 23.2  \\
\Approach~+ SoftGroup & 41.3 & 64.4 & 41.0 & 22.7 & 37.2 & 78.4 & 7.9 & 35.9 & 17.2 & 33.8 & 42.4 & 26.2 & 50.3 & 51.8 & 28.6 & 47.1 & 44.4 & 84.2 & 29.6 \\
\Approach~+ ISBNet & 50.6 & 76.3 & 45.5 & 28.5 & 46.0 & 82.7 & 21.8 & 41.3 & 22.0 & 51.3 & 51.3 & 55.9 & 44.5 & 52.8 & 59.7 & 49.5 & 52.8 & 90.2 & 39.5 \\
\Approach~+ SPFormer & 51.1 & 78.3 & 47.2 & 41.2 & 47.0 & 80.0 & 21.3 & 39.5 & 19.2 & 50.2 & 54.5 & 54.7 & 44.8 & 52.1 & 54.7 & 57.2 & 52.0 & 86.3 & 39.7 \\
\bottomrule
\end{tabular}
\caption{Per-class AP of 3D instance segmentation on the ScanNetV2 validation set. Our \Approach's versions of 3DIS methods achieve competitive performances with SOTA fully supervised version.}
\label{tab:scannet_val_detail}
\end{table*}

\begin{table*}[t]
\small
\setlength{\tabcolsep}{3pt}
\centering
\begin{tabular}{l|c|ccccccccccccccccccccc}
\toprule
Method & AP & \rotatebox{90}{bathtub} & \rotatebox{90}{bed} & \rotatebox{90}{bookshe.}      & \rotatebox{90}{cabinet}  & \rotatebox{90}{chair} & \rotatebox{90}{counter}  & \rotatebox{90}{curtain}  & \rotatebox{90}{desk}  & \rotatebox{90}{door}  & \rotatebox{90}{other}  & \rotatebox{90}{picture}  & \rotatebox{90}{fridge} & \rotatebox{90}{s.curtain}  & \rotatebox{90}{sink}  & \rotatebox{90}{sofa} & \rotatebox{90}{table} & \rotatebox{90}{toilet} & \rotatebox{90}{window} \\ 
\midrule
\rowcolor{gray!30} PointGroup \cite{jiang2020pointgroup}  & 40.7 & 63.9 & 49.6 & 41.5 & 24.3 & 64.5 & 2.1 & 57.0 & 11.4 & 21.1 & 35.9 & 21.7 & 42.8 & 66.0 & 25.6 & 56.2 & 34.1 & 86.0 & 29.1 \\
\rowcolor{gray!30} SSTNet \cite{liang2021instance}  & 50.6 & 73.8 & 54.9 & 49.7 & 31.6 & 69.3 & 17.8 & 37.7 & 19.8 & 33.0 & 46.3 & 57.6 & 51.5 & 85.7 & 49.4 & 63.7 & 45.7 & 94.3 & 29.0 \\
\rowcolor{gray!30} SoftGroup \cite{vu2022softgroup} & 50.4 & 66.7 & 57.9 & 37.2 & 38.1 & 69.4 & 7.2 & 67.7 & 30.3 & 38.7 & 53.1 & 31.9 & 58.2 & 75.4 & 31.8 & 64.3 & 49.2 & 90.7 & 38.8 \\
\rowcolor{gray!30} ISBNet \cite{ngo2023isbnet} & 55.9 & 92.6 & 59.7 & 39.0 & 43.6 & 72.2 & 27.6 & 55.6 & 38.0 & 45.0 & 50.5 & 58.3 & 73.0 & 57.5 & 45.5 & 60.3 & 57.3 & 97.9 & 33.2 \\ 
\rowcolor{gray!30} SPFormer \cite{sun2022superpoint} & 54.9 & 74.5 & 64.0 & 48.4 & 39.5 & 73.9 & 31.1 & 56.6 & 33.5 & 46.8 & 49.2 & 55.5 & 47.8 & 74.7 & 43.6 & 71.2 & 54.0 & 89.3 & 34.3 \\
\midrule
Box2Mask \cite{chibane2021box2mask} & 43.3 & 74.1 & 46.3 & 43.3 & 28.3 & 62.5 & 10.3 & 29.8 & 12.5 & 26.0 & 42.4 & 32.2 & 47.2 & 70.1 & 36.3 & 71.1 & 30.9 & 88.2 & 27.2 \\
\midrule
\Approach~+ PointGroup & 39.4 & 66.7 & 42.5 & 43.4 & 28.8 & 61.5 & 2.3 & 48.0 & 9.8 & 20.7 & 31.3 & 17.1 & 46.1 & 75.4 & 26.3 & 43.5 & 35.1 & 81.5 & 28.9  \\
\Approach~+ SSTNet & 45.8 & 85.2 & 47.2 & 38.2 & 35.8 & 66.7 & 13.1 & 37.2 & 19.0 & 32.3 & 40.8 & 28.3 & 34.3 & 86.3 & 43.1 & 52.6 & 46.3 & 92.9 & 24.5 \\
\Approach~+ SoftGroup & 42.1 & 55.3 & 45.6 & 35.7 & 28.9 & 69.0 & 4.3 & 47.1 & 19.7 & 29.3 & 37.5 & 19.4 & 51.9 & 71.8 & 24.8 & 57.4 & 44.3 & 86.9 & 28.8 \\
\Approach~+ ISBNet & 49.8 & 73.6 & 56.1 & 42.3 & 38.2 & 70.1 & 11.5 & 42.6 & 13.7 & 40.8 & 43.6 & 53.7 & 51.3 & 72.3 & 46.6 & 60.8 & 45.5 & 93.7 & 31.1  \\
\Approach~+ SPFormer & 48.2 & 73.9 & 50.5 & 44.2 & 38.1 & 71.5 & 6.4 & 41.4 & 18.9 & 43.0 & 43.5 & 55.0 & 35.4 & 70.7 & 42.7 & 67.6 & 43.5 & 84.2 & 36.7 \\
\bottomrule
\end{tabular}
\caption{Per-class AP of 3D instance segmentation on the ScanNetV2 hidden test set. Our \Approach's versions of 3DIS methods achieve competitive performances with SOTA fully supervised versions.}
\label{tab:scannet_test_detail}
\end{table*}

\subsection{More Qualitative Results of Our Approach}
\label{sec:more_qualitative} 

% \myheading{Quality of our pseudo labels on S3DIS dataset.} 
% We visualize the quality of the pseudo labels of Box2Mask \cite{chibane2021box2mask} and ours on the training set of S3DIS in Fig.~\ref{fig:quali_s3dis}.

\begin{figure*}[t]
  \centering
  \includegraphics[width=1.0\linewidth]{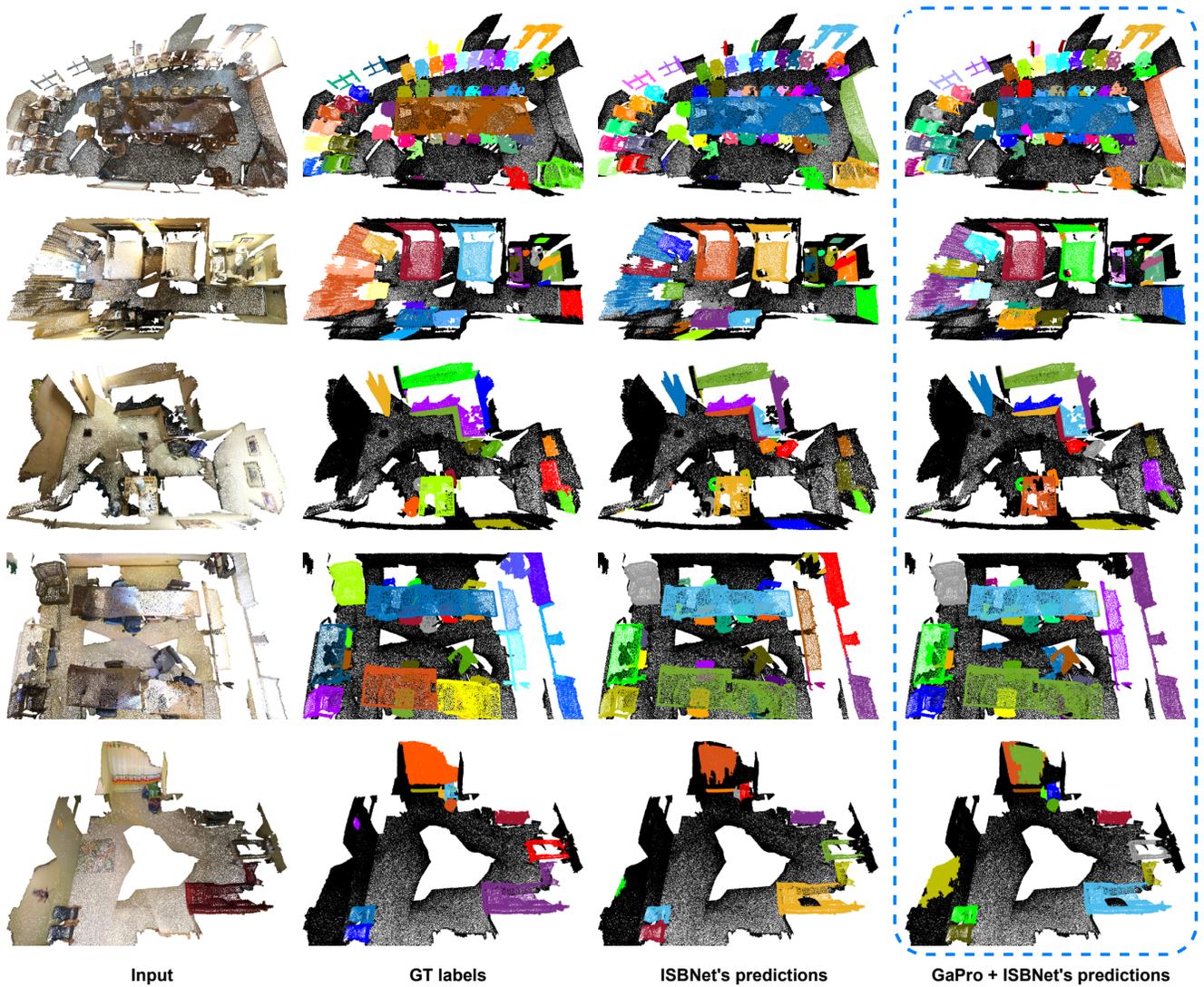}
\vspace{-10pt}
   \caption{The predicted instance masks of ISBNet \cite{ngo2023isbnet} trained with our pseudo labels and GT labels on ScanNetV2. Each row shows one example including the input point cloud, GT labels, predictions of ISBNet trained with GT labels, and predictions of ISBNet trained with our pseudo labels (\textcolor{blue}{dash box}). The ISBNet trained with our labels gives comparable results to the fully supervised counterpart, except for the last row.}
\vspace{-12pt}
   \label{fig:sup_more_quali_scannet}
\end{figure*}

\begin{figure*}[t]
  \centering
  \includegraphics[width=1.0\linewidth]{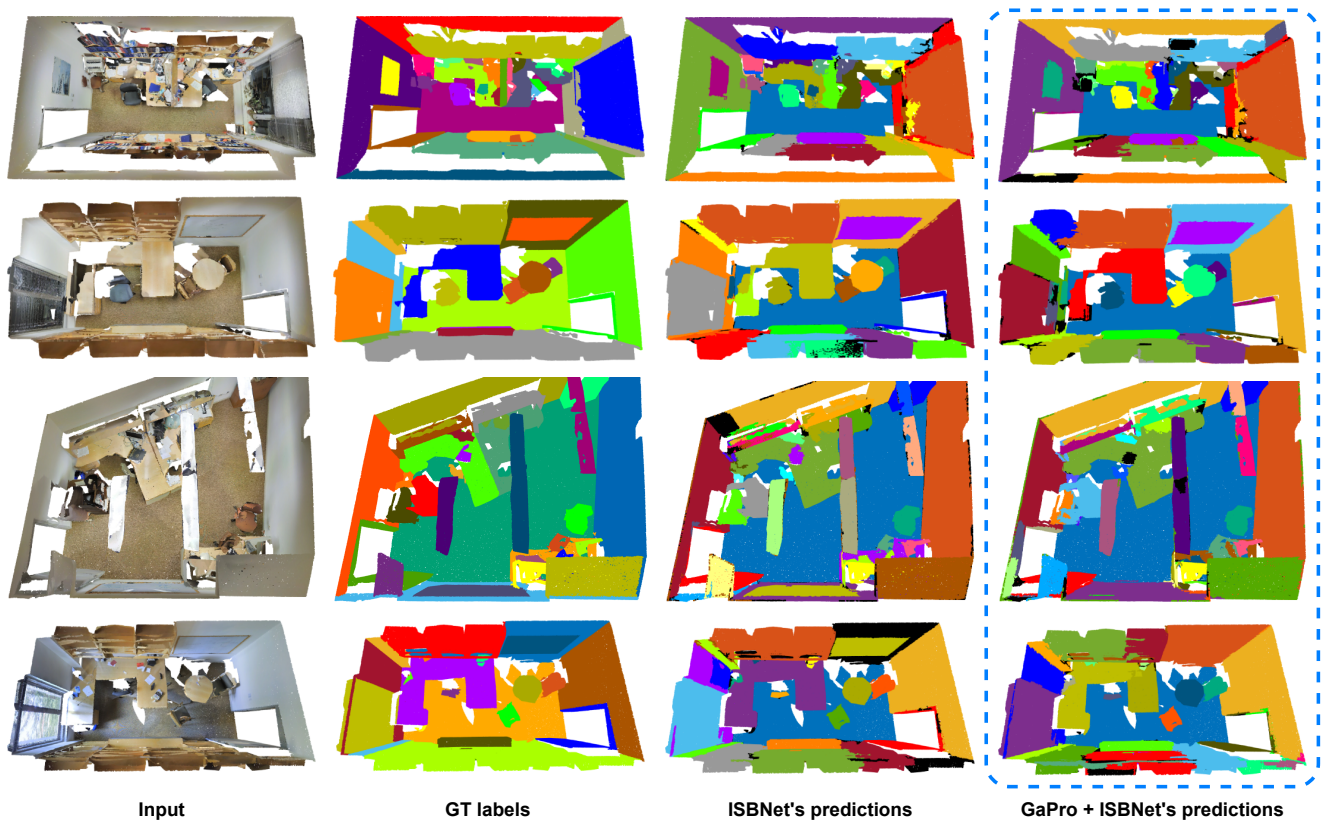}
\vspace{-10pt}
   \caption{The predicted instance masks of ISBNet \cite{ngo2023isbnet} trained with our pseudo labels and GT labels on S3DIS. Each row shows one example including the input point cloud, GT labels, predictions of ISBNet trained with GT labels, and predictions of ISBNet trained with our pseudo labels (\textcolor{blue}{dash box}). The ISBNet trained with our labels gives comparable results to the fully supervised counterpart, except for the last row}
\vspace{-12pt}
   \label{fig:sup_more_quali_s3dis}
\end{figure*}

% \myheading{Qualitative results of ISBNet trained with our pseudo labels and GT labels on ScanNetV2 and S3DIS datasets.}
The predicted instance masks of ISBNet \cite{ngo2023isbnet} trained with our pseudo labels and GT labels are visualized in \cref{fig:sup_more_quali_scannet} (for ScanNetV2) and \cref{fig:sup_more_quali_s3dis} (for S3DIS).

% \begin{figure*}[t]
%   \centering
%   \includegraphics[width=1.0\linewidth]{images/quali_s3dis.pdf}
% \vspace{-10pt}
%    \caption{\textbf{Quality of our pseudo labels compared with Box2Mask's and GT labels}. Each row from left to right shows an example with the input and axis-aligned bounding box labels, Box2Mask's pseudo labels, our pseudo labels, and GT labels. Our approach produces highly accurate instance masks, especially in regions with overlapping GT bounding boxes (\textcolor{blue}{blue} circles) where Box2Mask faces difficulty.}
% \vspace{-12pt}
%    \label{fig:quali_s3dis}
% \end{figure*}

\end{document}

%% file: definitions.tex
\def\mA{\mathcal{A}}
\def\mB{\mathcal{B}}
\def\mC{\mathcal{C}}
\def\mD{\mathcal{D}}
\def\mE{\mathcal{E}}
\def\mF{\mathcal{F}}
\def\mG{\mathcal{G}}
\def\mH{\mathcal{H}}
\def\mI{\mathcal{I}}
\def\mJ{\mathcal{J}}
\def\mK{\mathcal{K}}
\def\mL{\mathcal{L}}
\def\mM{\mathcal{M}}
\def\mN{\mathcal{N}}
\def\mO{\mathcal{O}}
\def\mP{\mathcal{P}}
\def\mQ{\mathcal{Q}}
\def\mR{\mathcal{R}}
\def\mS{\mathcal{S}}
\def\mT{\mathcal{T}}
\def\mU{\mathcal{U}}
\def\mV{\mathcal{V}}
\def\mW{\mathcal{W}}
\def\mX{\mathcal{X}}
\def\mY{\mathcal{Y}}
\def\mZ{\mathcal{Z}} 

\def\bbN{\mathbb{N}} 
\def\bbR{\mathbb{R}} 
\def\bbP{\mathbb{P}} 
\def\bbQ{\mathbb{Q}} 
\def\bbE{\mathbb{E}}

\def\1n{\mathbf{1}_n}
\def\0{\mathbf{0}}
\def\1{\mathbf{1}}

\def\A{{\bf A}}
\def\B{{\bf B}}
\def\C{{\bf C}}
\def\D{{\bf D}}
\def\E{{\bf E}}
\def\F{{\bf F}}
\def\G{{\bf G}}
\def\H{{\bf H}}
\def\I{{\bf I}}
\def\J{{\bf J}}
\def\K{{\bf K}}
\def\L{{\bf L}}
\def\M{{\bf M}}
\def\N{{\bf N}}
\def\O{{\bf O}}
\def\P{{\bf P}}
\def\Q{{\bf Q}}
\def\R{{\bf R}}
\def\S{{\bf S}}
\def\T{{\bf T}}
\def\U{{\bf U}}
\def\V{{\bf V}}
\def\W{{\bf W}}
\def\X{{\bf X}}
\def\Y{{\bf Y}}
\def\Z{{\bf Z}}

\def\a{{\bf a}}
\def\b{{\bf b}}
\def\c{{\bf c}}
\def\d{{\bf d}}
\def\e{{\bf e}}
\def\f{{\bf f}}
\def\g{{\bf g}}
\def\h{{\bf h}}
\def\i{{\bf i}}
\def\j{{\bf j}}
\def\k{{\bf k}}
\def\l{{\bf l}}
\def\m{{\bf m}}
\def\n{{\bf n}}
\def\o{{\bf o}}
\def\p{{\bf p}}
\def\q{{\bf q}}
\def\r{{\bf r}}
\def\s{{\bf s}}
\def\t{{\bf t}}
\def\u{{\bf u}}
\def\v{{\bf v}}
\def\w{{\bf w}}
\def\x{{\bf x}}
\def\y{{\bf y}}
\def\z{{\bf z}}

\def\balpha{\mbox{\boldmath{$\alpha$}}}
\def\bbeta{\mbox{\boldmath{$\beta$}}}
\def\bdelta{\mbox{\boldmath{$\delta$}}}
\def\bgamma{\mbox{\boldmath{$\gamma$}}}
\def\blambda{\mbox{\boldmath{$\lambda$}}}
\def\bsigma{\mbox{\boldmath{$\sigma$}}}
\def\btheta{\mbox{\boldmath{$\theta$}}}
\def\bomega{\mbox{\boldmath{$\omega$}}}
\def\bxi{\mbox{\boldmath{$\xi$}}}
\def\bnu{\mbox{\boldmath{$\nu$}}}                                  
\def\bphi{\mbox{\boldmath{$\phi$}}}
\def\bmu{\mbox{\boldmath{$\mu$}}}

\def\bDelta{\mbox{\boldmath{$\Delta$}}}
\def\bOmega{\mbox{\boldmath{$\Omega$}}}
\def\bPhi{\mbox{\boldmath{$\Phi$}}}
\def\bLambda{\mbox{\boldmath{$\Lambda$}}}
\def\bSigma{\mbox{\boldmath{$\Sigma$}}}
\def\bGamma{\mbox{\boldmath{$\Gamma$}}}
                                  
\newcommand{\myprob}[1]{\mathop{\mathbb{P}}_{#1}}

\newcommand{\myexp}[1]{\mathop{\mathbb{E}}_{#1}}

\newcommand{\mydelta}[1]{1_{#1}}

\newcommand{\myminimum}[1]{\mathop{\textrm{minimum}}_{#1}}
\newcommand{\mymaximum}[1]{\mathop{\textrm{maximum}}_{#1}}    
\newcommand{\mymin}[1]{\mathop{\textrm{minimize}}_{#1}}
\newcommand{\mymax}[1]{\mathop{\textrm{maximize}}_{#1}}
\newcommand{\mymins}[1]{\mathop{\textrm{min.}}_{#1}}
\newcommand{\mymaxs}[1]{\mathop{\textrm{max.}}_{#1}}  
\newcommand{\myargmin}[1]{\mathop{\textrm{argmin}}_{#1}} 
\newcommand{\myargmax}[1]{\mathop{\textrm{argmax}}_{#1}} 
\newcommand{\myst}{\textrm{s.t. }}

\newcommand{\denselist}{\itemsep -1pt}
\newcommand{\sparselist}{\itemsep 1pt}

\definecolor{pink}{rgb}{0.9,0.5,0.5}
\definecolor{purple}{rgb}{0.5, 0.4, 0.8}   
\definecolor{gray}{rgb}{0.3, 0.3, 0.3}
\definecolor{mygreen}{rgb}{0.2, 0.6, 0.2}

\newcommand{\cyan}[1]{\textcolor{cyan}{#1}}
\newcommand{\red}[1]{\textcolor{red}{#1}}  
\newcommand{\blue}[1]{\textcolor{blue}{#1}}
\newcommand{\magenta}[1]{\textcolor{magenta}{#1}}
\newcommand{\pink}[1]{\textcolor{pink}{#1}}
\newcommand{\green}[1]{\textcolor{green}{#1}} 
\newcommand{\gray}[1]{\textcolor{gray}{#1}}    
\newcommand{\mygreen}[1]{\textcolor{mygreen}{#1}}    
\newcommand{\purple}[1]{\textcolor{purple}{#1}}       

\definecolor{greena}{rgb}{0.4, 0.5, 0.1}
\newcommand{\greena}[1]{\textcolor{greena}{#1}}

\definecolor{bluea}{rgb}{0, 0.4, 0.6}
\newcommand{\bluea}[1]{\textcolor{bluea}{#1}}
\definecolor{reda}{rgb}{0.6, 0.2, 0.1}
\newcommand{\reda}[1]{\textcolor{reda}{#1}}

\def\changemargin#1#2{\list{}{\rightmargin#2\leftmargin#1}\item[]}
\let\endchangemargin=\endlist
                                               
\newcommand{\cm}[1]{}

\newcommand{\mhoai}[1]{{\color{magenta}\textbf{[MH: #1]}}}

\newcommand{\mtodo}[1]{{\color{red}$\blacksquare$\textbf{[TODO: #1]}}}
\newcommand{\myheading}[1]{\vspace{1ex}\noindent \textbf{#1}}
\newcommand{\htimesw}[2]{\mbox{$#1$$\times$$#2$}}

% The following are useful for creating homework or exams

\newif\ifshowsolution
%\showsolutionfalse
\showsolutiontrue

\ifshowsolution  
\newcommand{\Comment}[1]{\paragraph{\bf $\bigstar $ COMMENT:} {\sf #1} \bigskip}
\newcommand{\Solution}[2]{\paragraph{\bf $\bigstar $ SOLUTION:} {\sf #2} }
\newcommand{\Mistake}[2]{\paragraph{\bf $\blacksquare$ COMMON MISTAKE #1:} {\sf #2} \bigskip}
\else
\newcommand{\Solution}[2]{\vspace{#1}}
\fi

\newcommand{\truefalse}{
\begin{enumerate}
	\item True
	\item False
\end{enumerate}
}

\newcommand{\yesno}{
\begin{enumerate}
	\item Yes
	\item No
\end{enumerate}
}

\newcommand{\Sref}[1]{Sec.~\ref{#1}}
\newcommand{\Eref}[1]{Eq.~(\ref{#1})}
\newcommand{\Fref}[1]{Fig.~\ref{#1}}
\newcommand{\Tref}[1]{Table~\ref{#1}}